%% file: main.tex
\documentclass[11pt]{article}

\usepackage[final]{acl}

\usepackage{times}
\usepackage{latexsym}
\usepackage[T1]{fontenc}
\usepackage[utf8]{inputenc}
\usepackage{microtype}
\usepackage[final]{graphicx}
\graphicspath{{./}{paper/}}
\usepackage{amsmath}
\usepackage{amssymb}
\usepackage{booktabs}
\usepackage{url}
\usepackage{xcolor}
\usepackage{float}
\usepackage{enumitem}
\usepackage{tabularx}
\usepackage{multirow}
\usepackage{algorithm}
\usepackage{algpseudocode}
\usepackage{threeparttable}
\usepackage[most]{tcolorbox}
\tcbuselibrary{breakable, listings}
\usepackage{listings}
\usepackage{subcaption}
\setlist[itemize]{leftmargin=*,topsep=2pt,itemsep=1pt,parsep=0pt,partopsep=0pt}
\newtcolorbox{promptbox}[1][] {
    breakable,
    enhanced,
    title=#1,
    colback=white,
    colframe=black!70!white,
    width=\linewidth,
    fonttitle=\bfseries\footnotesize,
    fontupper=\footnotesize,
    arc=1mm,
    boxrule=0.5pt,
    left=4pt,
    right=4pt,
    top=4pt,
    bottom=4pt,
    before skip=10pt,
    after skip=10pt,
    enlarge left by=0mm,
    boxsep=0pt
}

\definecolor{demph}{gray}{0.6}

\title{SurveyGraph: Generating Academic Surveys via\\
Methods-Entity Knowledge Graphs}

\author{
Yifan Yuan \quad Yancheng Zhu \quad Zuoyu Zhang \quad Xiaogang Peng\thanks{Corresponding author.} \\
School of Artificial Intelligence, Shenzhen University, Shenzhen, China \\
\texttt{yifanyuan@szu.edu.cn}, \texttt{2410673009@mails.szu.edu.cn} \\
\texttt{2310533003@email.szu.edu.cn}, \texttt{pengxg@szu.edu.cn} \\
}

\begin{document}
\maketitle

\begin{abstract}
Automated survey generation systems increasingly rely on flat retrieval over document collections, limiting their ability to capture the evolutionary structure of research fields. We present \textsc{SurveyGraph}, a framework centered on a \textsc{Methods-Entity Knowledge Graph (MEKG)} that encodes method-level follow-on, competition, and shared-resource relations across papers. The MEKG serves as a unified backbone for core-paper discovery, graph-aware retrieval, hybrid outline generation, and evolutionary narrative synthesis.

We evaluate SurveyGraph through a multi-level protocol spanning retrieval benchmarks, semantic-edge validation, citation grounding audits, benchmark-facing collection probes, and same-backend generation-stage diagnostics. In a controlled seven-method equal-budget benchmark, MEKG-anchored fusion achieves the best balance of recall and structural density among all tested methods, demonstrating that typed graph structure provides a complementary signal to simpler retrieval approaches. On the archived graph-native core diagnostic, graph-aware retrieval nearly doubles recall over flat lexical baselines across all eight runs ($p{=}0.004$). A semantic evidence gate with AUC=0.75 filters noisy edges, and citation-grounding audits over 160 generated sentences confirm 80.6\% anchored-or-partial support with evidence-level contrasts validating genuine alignment. Human validation over 108 stratified semantic edges (Cohen's $\kappa{=}0.64$ for the binary support decision) confirms 68.5\% supported-or-partial after adjudication, with \texttt{improves} relations strongest at 80.6\% and \texttt{extends} weakest at 52.8\%. A SurveyBench-facing collection probe reaches 0.521 mean reference coverage on four topics under an eight-paper budget; same-backend Qwen diagnostics over a 30-paper/74-relation MEKG further show that removing hybrid outlining lowers artifact organization signals. We report concurrent-system comparisons only as bounded diagnostics rather than same-backend end-to-end claims. Code and artifacts are available at: \url{https://github.com/xzzvsxd/SurveyGraph}.
\end{abstract}
\section{Introduction}
\label{sec:introduction}

Scientific publishing continues to accelerate~\citep{bornmann2015growth}, making coherent, well-grounded survey writing increasingly expensive and time-consuming. This pressure has motivated a fast-growing line of LLM-based survey generation systems, including \texttt{AutoSurvey} \citep{wang2024autosurvey}, \texttt{SurveyX} \citep{liang2025surveyx}, \texttt{SurveyForge} \citep{chi2025surveyforge}, and \texttt{SciSage} \citep{shi2025scisage}. These systems have improved literature collection, outline drafting, and long-form generation, but most still operate over a document-centric view of the literature.

The core bottleneck is \emph{structural blindness}. Flat RAG pipelines retrieve papers by lexical or embedding similarity, yet high-quality survey writing depends on recovering the \emph{evolutionary structure} of a field: which work extends which idea, which methods branch into competing families, and which benchmarks define phase transitions. For example, a survey of LLM-based agents should distinguish ReAct-style reasoning--action loops \citep{yao2022react} from Toolformer-style self-supervised tool use \citep{schick2023toolformer}, rather than treating these papers as interchangeable topical matches. Without explicit relations, generated surveys become a collage of local summaries rather than a structured account of scientific development.

The automated survey-generation landscape changed rapidly in 2025--2026. New systems increasingly inject planning, memory, and structure into the generation loop, including evolving plans in \texttt{SurveyGen-I} \citep{chen2025surveygeni}, quality-aware retrieval in \texttt{SurveyGen} \citep{bao2025surveygen}, multi-view structure-first generation in \texttt{MVSS} \citep{liu2026mvss}, and structured retrieval in \texttt{StructSurvey} \citep{pedinotti2026structsurvey}. In parallel, evaluation has become stricter, with benchmarks such as \texttt{SurveyBench} \citep{sun2025surveybench}, \texttt{SurveyLens} \citep{guo2026surveylens}, and \texttt{DeepSurvey-Bench} \citep{zhang2026deepsurveybench}. This landscape demands systems that not only generate high-quality text but also provide transparent, verifiable evidence for their structural decisions.

To address this, we propose \textsc{SurveyGraph}, a framework centered on a \textsc{Methods-Entity Knowledge Graph (MEKG)} that encodes method-level follow-on, competition, shared tasks, and temporal dependencies. The graph is reused throughout the pipeline: it supports core-paper discovery, hybrid outline generation, graph-aware retrieval, and narrative synthesis. Our evaluation combines controlled retrieval benchmarks, semantic-edge validation, citation grounding audits, and SurveyBench-facing collection probes to demonstrate the value and limits of typed graph structure.

Our contributions are threefold. First, we present \textsc{SurveyGraph}, a graph-centric framework coupling MEKG construction, graph-aware retrieval, hybrid outline generation, and evolutionary narrative synthesis; unlike flat-RAG or citation-only approaches, MEKG encodes typed method-level relations that enable auditable, structure-preserving retrieval and generation. Second, we conduct a multi-level evaluation: (i)~a seven-method equal-budget retrieval benchmark demonstrating that MEKG-anchored fusion achieves the best recall--density tradeoff, (ii)~statistically significant gains on the archived graph-native diagnostic ($p{=}0.004$), (iii)~citation-grounding audits confirming 80.6\% anchored-or-partial support, (iv)~human validation ($\kappa{=}0.64$, 108 edges), (v)~SurveyBench-facing collection and bounded comparison diagnostics, and (vi)~same-backend Qwen generation diagnostics that clarify the role of hybrid outlining/evolution modules without claiming leaderboard parity. Third, we release all artifacts, evaluation scripts, the annotation package, and reproducibility traces.

\section{Related Work}
\label{sec:related_work}

\textit{Automated survey generation.}
Recent systems have moved from one-shot summarization toward multi-stage pipelines. \texttt{AutoSurvey} \citep{wang2024autosurvey} established the retrieval--outline--draft paradigm; \texttt{SurveyX} \citep{liang2025surveyx} and \texttt{SurveyForge} \citep{chi2025surveyforge} explored stronger preprocessing, outline heuristics, and re-polishing; \texttt{SciSage} \citep{shi2025scisage} and \texttt{ARISE} \citep{wang2025arise} introduced multi-agent and rubric-guided iterative refinement. More recent methods such as \texttt{SurveyGen-I} \citep{chen2025surveygeni} and \texttt{SurveyGen} \citep{bao2025surveygen} further emphasize evolving plans and quality-aware retrieval.

\textit{Structure-first and agentic extensions.}
Several concurrent systems target structural organization. \texttt{InteractiveSurvey} \citep{wen2025interactivesurvey} supports personalized and interactive survey construction; agentic variants such as \texttt{Agentic AutoSurvey} \citep{liu2025agenticautosurvey} and \texttt{LLM$\times$MapReduce-V2/V3} \citep{wang2025llmxmapreducev2,chao2025llmxmapreducev3} emphasize specialized orchestration. Among the closest concurrent works, \texttt{MVSS} \citep{liu2026mvss} aligns citation-grounded trees with survey text, while \texttt{StructSurvey} \citep{pedinotti2026structsurvey} pushes structural burden into retrieval.

\textit{Benchmarks and evaluation.}
\texttt{SurGE} \citep{su2025surge}, \texttt{SGSimEval} \citep{guo2025sgsimeval}, \texttt{SurveyBench} \citep{sun2025surveybench}, \texttt{SurveyLens} \citep{guo2026surveylens}, and \texttt{DeepSurvey-Bench} \citep{zhang2026deepsurveybench} increasingly target comprehensiveness, citation accuracy, and scholarly value. Complementary work such as \texttt{ReportBench} \citep{li2025reportbench}, \texttt{OpenScholar} \citep{asai2024openscholar}, and iterative pairwise evaluation settings reinforce the importance of citation grounding.

\textit{Graph-RAG and positioning.}
Recent graph-centric scholarly tools emphasize domain-specific literature subgraph curation and benchmarking \citep{varvarigos2026litbench}, document-level KG construction \citep{zhang2025rakg}, practical GraphRAG pipelines \citep{min2025practicalgraphrag}, and graph-guided concept selection \citep{liu2025graphguidedconcept}. SurveyGraph differs in two ways: (1) its central representation is a \emph{method--entity evolutionary graph} rather than a generic citation graph---enabling typed, auditable retrieval decisions; (2) the graph serves the entire pipeline from retrieval through outline to generation, creating a unified structural backbone.

\section{The SurveyGraph Framework}
\label{sec:methodology}

\begin{figure*}[t!]
    \centering
    \includegraphics[width=\textwidth]{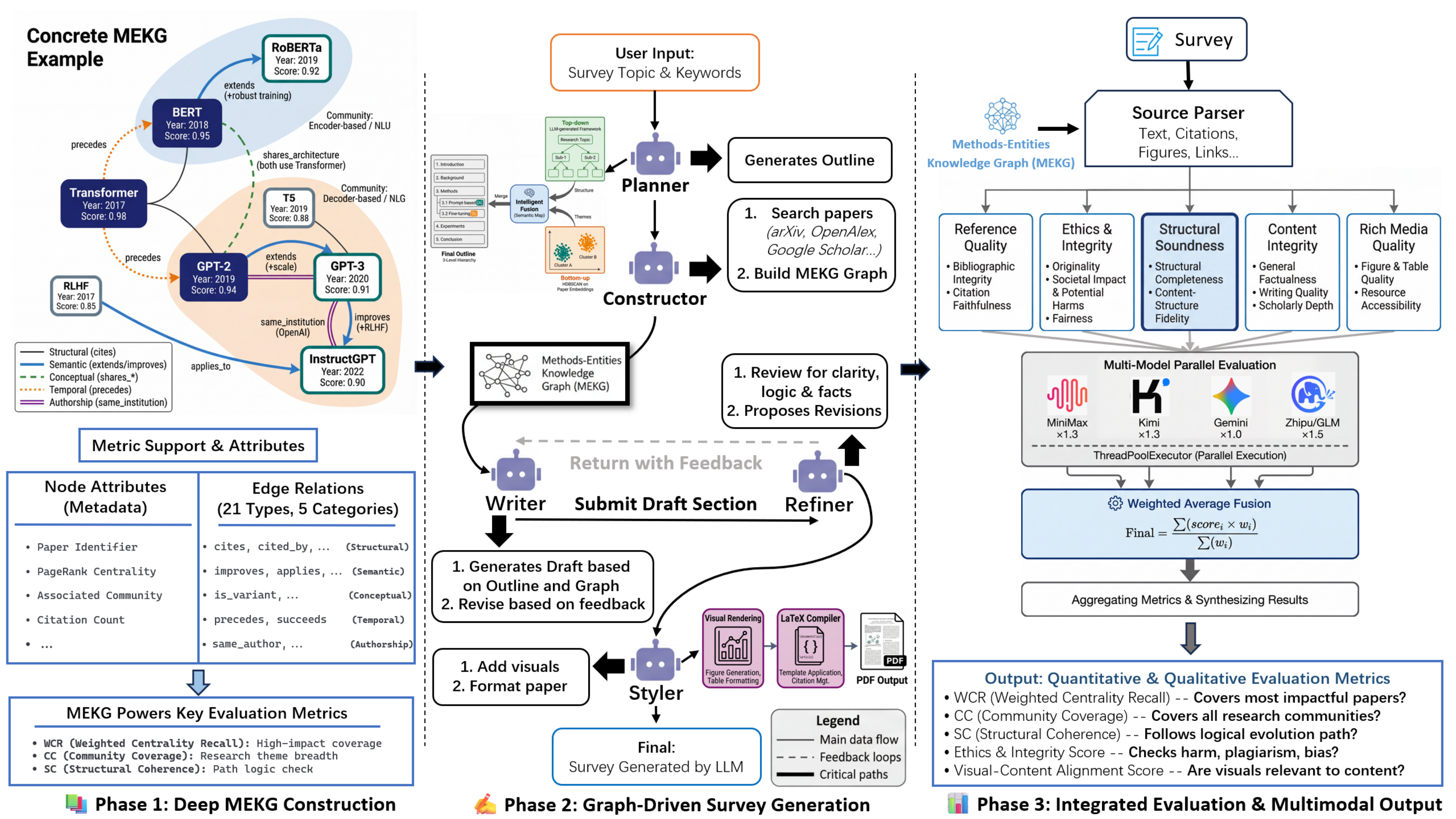}
    \caption{The \textsc{SurveyGraph} pipeline. Starting from a user-specified topic, the system constructs a Methods-Entity Knowledge Graph through multi-source collection and four-layer relation inference, then uses the graph for core-paper selection, hybrid outline generation, graph-aware retrieval, and evolutionary narrative synthesis.}
    \label{fig:pipeline}
\end{figure*}

\textsc{SurveyGraph} operates as a multi-stage pipeline designed to transform discrete paper data into coherent scientific narratives. Our methodology is structured into four stages: (1) \textit{MEKG Construction} to surface field-level structure; (2) \textit{Hybrid Outline Generation} to balance structural rigor with thematic discovery; (3) \textit{Graph-RAG Retrieval} to retrieve graph-linked context; and (4) \textit{Evolutionary Narrative Synthesis} to transform graph signals into a coherent scholarly history. Figure~\ref{fig:pipeline} summarizes the end-to-end workflow.

\subsection{Stage 1: MEKG Construction}
\label{sec:mekg_construction}

To build an operational graph approximation of method-level structure, we construct the \textsc{Methods-Entity Knowledge Graph (MEKG)}. MEKG is formally defined as $G = (V, E)$, where $V$ contains heterogeneous entities (\texttt{Method}, \texttt{Task}, \texttt{Dataset}, \texttt{Metric}, \texttt{Model}, \texttt{Innovation}) and $E$ comprises 21 typed relations across five categories, summarized in Table~\ref{tab:relation_types}. Table~\ref{tab:relation_types} describes the intended schema; the verified archived runs analyzed in this paper expose only a subset of these relation families in saved artifacts, so all empirical retrieval analyses below are restricted to relation types actually present in the local outputs.

We distinguish between key semantic relations to surface follow-on links between \emph{paper records}. Specifically, \texttt{extends} denotes a methodological follow-on relation, \texttt{improves} denotes a performance-oriented refinement, and \texttt{applies} denotes a transfer of a method or paradigm to a new task/domain. These labels are inferred from paper metadata, extracted text, and relation evidence through a multi-layer inference engine; they are validated through the semantic-edge protocol described in Section~\ref{sec:experiments}. For entity resolution, we leverage scholarly metadata for papers and authors while resolving methodological entities that are often absent from generic academic knowledge graphs. In the repository implementation, upstream paper merging also applies a lightweight title normalization pass (lowercasing, punctuation stripping, and whitespace collapsing) before title-hash deduplication, so obvious cross-source aliases are filtered before MEKG construction.

Our pipeline constructs the MEKG via a four-layer hybrid inference engine. \textit{Structural Verification} establishes the graph backbone using bibliometric data; \textit{Contextual Pattern Matching} identifies explicit relations in citation contexts; \textit{Shared Entity Analysis} links conceptually related papers through extracted entities; and \textit{LLM Deep Reasoning} proposes implicit relationships that require broader scholarly context. The relation prompt returns a typed edge, confidence, and rationale from two paper records, and the design allows a \texttt{none} outcome when only topical similarity is evident; Appendix~\ref{app:prompts} gives the template and Appendix~\ref{app:diagnostic_additions} reports validation audits. Semantic edges are thus treated downstream as confidence-weighted links that can be filtered through the evidence gate described in Section~\ref{sec:experiments}.

To improve robustness across models with different context windows, the implementation adaptively selects between direct full-text analysis and a \textit{Recursive Snowball Chunking} strategy that processes overlapping segments while carrying forward structured summaries.

\begin{table}[t]
\centering
\footnotesize
\setlength{\tabcolsep}{3pt}
\begin{tabularx}{\columnwidth}{l >{\raggedright\arraybackslash}X ccc}
\toprule
\textbf{Cat.} & \textbf{Relations} & \textbf{Layer} & \textbf{Conf.} & \textbf{\#} \\
\midrule
Struct. & cites, cited\_by, co\_cite, bib\_coup & 1 & 1.0 & 4 \\
Semantic & extends, improves, applies, ... & 4 & $\geq$0.6 & 8 \\
Concept. & shares\_*, is\_variant/instance\_of & 2--3 & 0.7--0.9 & 5 \\
Temporal & precedes, succeeds & 1 & 1.0 & 2 \\
Author. & same\_author, same\_group & 1 & 1.0 & 2 \\
\midrule
\textbf{Total} & & & & \textbf{21} \\
\bottomrule
\end{tabularx}
\caption{MEKG relation schema with inference layer and confidence thresholds. Layer 1 is rule-based, Layers 2--3 use pattern/entity matching, and Layer 4 uses LLM reasoning.}
\label{tab:relation_types}
\end{table}

\begin{figure}[t]
    \centering
    \includegraphics[width=\columnwidth]{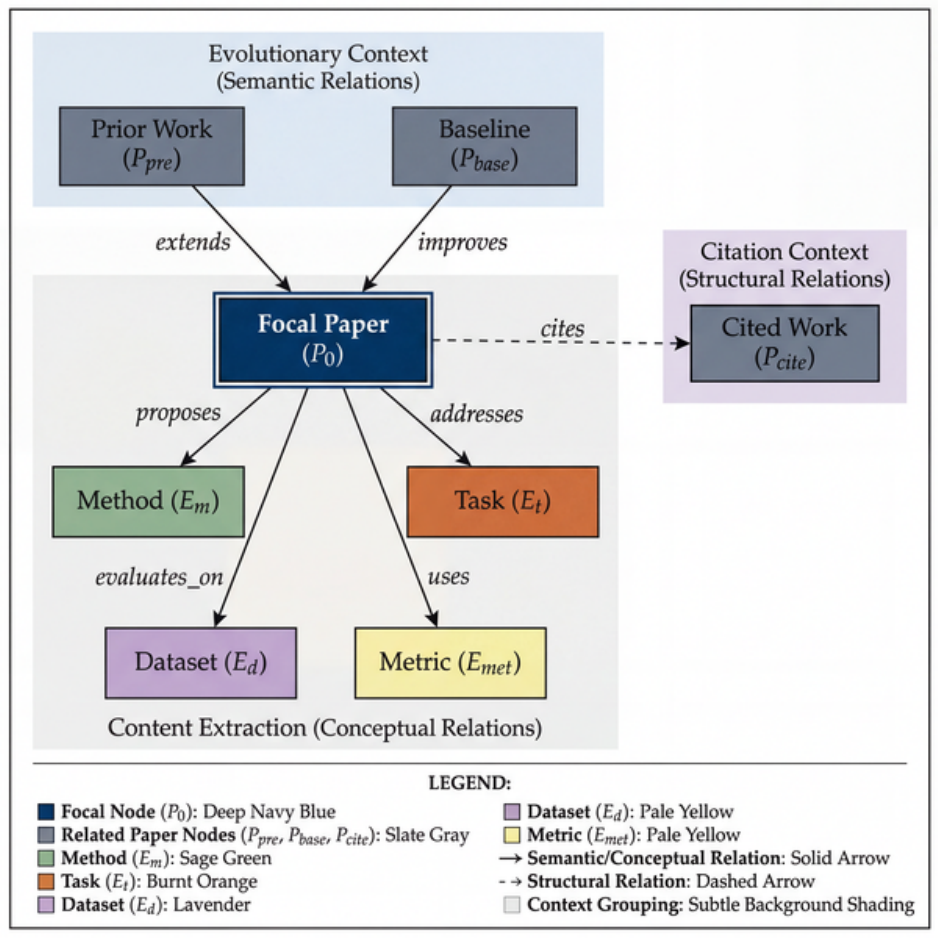}
    \caption{MEKG schema overview. Heterogeneous entity types (Method, Task, Dataset, Metric, Model, Innovation) are connected through 21 typed relations across five categories, enabling structured representation of method-level evolutionary relationships.}
    \label{fig:mekg_schema}
\end{figure}

To identify influential works, we use a weighted PageRank-style score for \emph{core paper discovery}:
\begin{equation}
\begin{aligned}
S(p_i) &= \alpha \cdot \text{PR}(p_i) + \beta \cdot \log(1 + C_i) \\
       &\quad + \gamma \cdot R_i + \delta \cdot D_i
\end{aligned}
\end{equation}
where PR is the PageRank score on the available paper--paper relation graph, $C_i$ is citation count, $R_i = 1/(Y_{\text{curr}} - Y_i + 1)$ is the recency boost, and $D_i$ is degree centrality. In the current implementation, we use fixed hand-tuned weights $\alpha=0.4$, $\beta=0.3$, $\gamma=0.2$, and $\delta=0.1$ as defaults rather than learned parameters, and an adaptive elbow heuristic then selects the highest-scoring papers as the graph's ``core'' set. A small archived-run sensitivity audit in Appendix~\ref{app:diagnostic_additions} shows that the resulting top-$10$ core-paper set is fairly stable to moderate reweighting, and an OpenAlex title-matched citation-proxy replay recomputes the same score under recovered bibliometric signal.

\subsection{Stage 2: Hybrid Outline Generation}
\label{sec:hybrid_outline}

The goal of this stage is to establish a survey structure that is both academically rigorous and grounded in the actual data distribution of the field. We fuse a top-down survey template with bottom-up themes discovered through HDBSCAN clustering on paper embeddings. The current implementation uses title+abstract embeddings from the HuggingFace fallback stack, L2-normalizes them, clusters with HDBSCAN using $\texttt{min\_cluster\_size}=3$ and $\texttt{min\_samples}=2$ (both adaptively raised with sample count), and falls back to KMeans when HDBSCAN is unavailable; an LLM-based structural architect then merges the clusters into a three-level hierarchy (section $\rightarrow$ subsection $\rightarrow$ subsubsection).

\subsection{Stage 3: Graph-RAG Retrieval}
\label{sec:graph_rag}

To overcome the structural blindness of flat RAG, Graph-RAG traverses the MEKG topology to retrieve graph-linked context for each section. For a target section, the system identifies core papers and expands the context window through predecessor, successor, and sibling relations. The current implementation uses a 1-hop traversal strategy with a retrieval budget of $k=5$ papers per relation type, ranked by a hybrid score of MEKG centrality, category/subcategory overlap, and the same title+abstract embedding backbone used for clustering.

Context windows are populated with topologically relevant papers: \textbf{Predecessors} (papers the current work \texttt{extends}/\texttt{improves}), \textbf{Successors} (papers that build upon the current work), and \textbf{Siblings} (papers sharing \texttt{method}/\texttt{task}/\texttt{dataset}). Our post-hoc depth sweep in Appendix~\ref{app:diagnostic_additions} shows that 2-hop expansion increases relation density but lowers community coverage and unique-paper yield while traversing far more edges, so the current default remains 1-hop with adaptive pruning when the context window grows too large.

\subsection{Stage 4: Evolutionary Narrative Synthesis}
\label{sec:evolutionary}

The final stage transforms structured graph signals and retrieved contexts into evolution-aware prose. An evolutionary narrative analyzer identifies likely milestones and phase transitions, then organizes them into a chronological and logical survey narrative. The implementation follows a five-stage process: (1) statistics over temporal distribution, (2) milestone identification, (3) phase analysis, (4) trend analysis, and (5) final narrative generation.

Each section undergoes a multi-stage refinement process using write--evaluate--refine loops (Algorithm~\ref{alg:reflection}). A section-level reflection agent assesses coverage and coherence. For rich media, the system uses \emph{data-template separation}: LLMs extract structured JSON-like data, which is rendered with pre-defined TikZ/\LaTeX{} templates. This design improves render-time validation and reduces prompt fragility compared with directly asking the generator to emit finished figures.

\begin{algorithm}[t]
\caption{Section-Level Reflection}
\label{alg:reflection}
\small
\begin{algorithmic}[1]
\Require Section $s$, threshold $\tau$=3.5, max iterations $K$=3
\State $content \gets \text{Generate}(s)$
\For{$k = 1$ to $K$}
    \State $score, feedback \gets \text{Evaluate}(content)$ \Comment{Coverage, coherence, citations, critical analysis}
    \If{$score \geq \tau$}
        \textbf{break}
    \EndIf
    \State $content \gets \text{Refine}(content, feedback)$
\EndFor
\State \Return $content$
\end{algorithmic}
\end{algorithm}

\section{Evaluation Framework}
\label{sec:evaluation}

We design a multi-level evaluation framework that separates graph-native diagnostics, retrieval benchmarks, grounding audits, and benchmark alignment.

\subsection{Graph-Native Metrics}
\label{sec:diagnostic_metrics}

We define three graph-native metrics to measure how strongly a generated survey tracks the structure of its underlying MEKG. \textit{Weighted Centrality Recall (WCR)} measures the coverage of core papers weighted by their importance $S(p_i)$:
\[
\text{WCR} = \frac{\sum_{p \in \mathcal{P}_{\text{cited}} \cap \mathcal{P}_{\text{core}}} S(p)}{\sum_{p \in \mathcal{P}_{\text{core}}} S(p)}.
\]
\textit{Community Coverage (CC)} measures how many graph communities are represented. \textit{Structural Coherence (SC)} is the fraction of communities matched by at least one outline section title.

\subsection{Benchmark Alignment}
\label{sec:benchmark_alignment}

We align our evaluation with metrics emphasized by \texttt{SurGE} \citep{su2025surge}, \texttt{SurveyBench} \citep{sun2025surveybench}, \texttt{SurveyLens} \citep{guo2026surveylens}, \texttt{DeepSurvey-Bench} \citep{zhang2026deepsurveybench}, and \texttt{ReportBench} \citep{li2025reportbench}. Our citation grounding audits address citation support and section--reference alignment, two dimensions represented across these benchmarks. The repository's internal evaluator (Appendix~\ref{app:current_eval_impl}) provides additional quality assurance across five dimensions.

\section{Evaluation}
\label{sec:experiments}

We organize our evaluation around five research questions: \textit{RQ1}: Does MEKG-based retrieval provide measurable advantages over flat and graph-free baselines? \textit{RQ2}: How does typed MEKG retrieval compare against lightweight graph alternatives in a controlled equal-budget benchmark? \textit{RQ3}: Can semantic-edge quality be validated, and does evidence-based filtering improve retrieval? \textit{RQ4}: Are generated survey citations well-grounded in source evidence? \textit{RQ5}: Under a fixed backend and reused MEKG substrate, how do generation-stage controls affect inspectable survey artifacts?

\subsection{Evaluation Setup}

Our evaluation covers eight completed SurveyGraph runs spanning seven topic labels: LLM function calling, document visual question answering, memory in AI agents, city-airspace embodied QA (two independent runs), Text-to-SQL agents, document question answering, and general agents. A focused same-backend generation-diagnostic batch uses LLM function calling. We conduct a seven-method equal-budget retrieval benchmark (Table~\ref{tab:main_benchmark}), semantic-edge validation with adjudicated human annotation, citation-grounding audits at claim and section levels, four artifact-backed Qwen component diagnostics, and one matched flat generation run on the same 30-paper corpus. Retrieval methods share identical section queries and top-$5$ budgets. Appendix~\ref{app:artifacts} lists every archived run; Appendix~\ref{app:generation_diagnostics} maps the generation diagnostics to their artifacts.

\subsection{Graph-Aware Retrieval (RQ1)}

We compare graph-aware retrieval (incorporating category overlap, PageRank centrality, and 1-hop relation expansion) against flat lexical retrieval across eight runs under a fixed top-$5$ per-section budget. Each run constructs a MEKG with 2,136--2,504 relations over 143 papers on average.

On the archived graph-native target set, graph-aware retrieval nearly doubles core-paper recall (0.373 vs.\ 0.216) with consistent gains across all runs (sign test $p{=}0.004$). Relation density improves substantially (0.259 vs.\ 0.162), and community coverage rises from 0.758 to 0.863. A signal-component ablation confirms that graph-neighborhood traversal is the primary driver: removing 1-hop expansion collapses recall to 0.219, while removing PageRank centrality has a smaller effect (recall drops to 0.341). We therefore treat this result as a mechanism diagnostic; the external-proxy benchmark below tests transfer beyond graph-derived targets.

The gain is robust to confidence thresholds: retaining only high-confidence edges ($\geq$0.9, representing 39\% of all edges) still yields recall of 0.342, indicating that the most confident edges carry most of the retrieval signal. Budget sweeps at top-3/5/7 all show consistent advantages.

To reduce circularity, we also evaluate against an external proxy built from papers cited in local survey documents but not selected by the MEKG core scorer. In a separate top-$20$ leave-one-run-out core-score calibration, external-proxy recall reaches 0.289; this setting is distinct from the top-$5$ equal-budget benchmark below, where typed MEKG alone reaches 0.153. The calibration remains topic-sensitive and is evidence that the score can be tuned against an external signal, not a claim that typed retrieval maximizes raw recall.

\subsection{Equal-Budget Retrieval Benchmark (RQ2)}

To contextualize MEKG against strong baselines, we conduct a seven-method benchmark under identical conditions (Table~\ref{tab:main_benchmark}). The ``MEKG typed'' row isolates typed-edge traversal alone (without the full graph-aware scoring from RQ1 that adds PageRank and 1-hop expansion); fusion rows show how combining both signals recovers and exceeds flat-retrieval recall.

\begin{table}[t]
\centering
\small
\setlength{\tabcolsep}{2.5pt}
\begin{tabular}{@{}lccccc@{}}
\toprule
\textbf{Method} & \textbf{R@5} & \textbf{P@5} & \textbf{Str.} & \textbf{Den.} & \textbf{CoCit} \\
\midrule
Flat lexical & .252 & .132 & .558 & .134 & .089 \\
Biblio.\ RRF & .309 & .194 & .530 & .132 & .142 \\
TF-IDF GraphRAG & .347 & .176 & .485 & .140 & .118 \\
MEKG typed & .153 & .089 & .392 & \textbf{.228} & .156 \\
Safe fusion & .261 & .134 & .528 & .179 & .147 \\
Biblio+TF-IDF & \textbf{.357} & \textbf{.200} & .498 & .134 & .131 \\
Biblio+MEKG & .304 & .178 & .486 & .174 & \textbf{.163} \\
\bottomrule
\end{tabular}
\caption{Seven-method equal-budget retrieval benchmark (top-$5$). \textit{Str.}=section-evidence alignment; \textit{Den.}=typed relation density; \textit{CoCit}=co-citation connectivity (fraction of retrieved pairs sharing $\geq$1 common citing paper via OpenAlex, independent of MEKG).}
\label{tab:main_benchmark}
\end{table}

The results reveal a complementarity pattern. Graph-free methods achieve higher raw recall but minimal typed structure (density $\leq$0.140). MEKG retrieval yields the densest neighborhoods but lower surface recall alone. \emph{Fusion methods outperform either in isolation}: MEKG-anchored fusion nearly doubles relation density while maintaining competitive recall. The MEKG-independent co-citation metric (CoCit) confirms that MEKG-linked methods produce retrieved sets with higher co-citation connectivity (0.147--0.163 vs.\ 0.089--0.131), indicating genuine bibliometric relationships rather than artifacts of MEKG's own edge definitions. An encoder-sensitivity analysis across five rankers (BM25, Sentence-BERT \citep{reimers2019sentencebert}, BGE, GTE, and E5) confirms that anchor-preserving fusion is the robust default: all encoders show the same complementarity pattern, with fusion consistently outperforming either signal alone regardless of the underlying embedding model.

\subsection{Semantic-Edge Validation (RQ3)}

We validate MEKG semantic edges through consistency audits, evidence-gated retrieval, and adjudicated human annotation.

\noindent\textbf{Consistency and negative controls.} Across 2,091 semantic edges (\texttt{extends}: 1,276; \texttt{improves}: 366; \texttt{applies}: 449), temporal direction consistency exceeds 98\% for all types. A deterministic evidence protocol examines lexical overlap between source/target paper metadata and labels each edge as supported, partial, or unsupported. Overall, 58.8\% are supported-or-partial, with \texttt{improves} strongest (80.6\%), \texttt{applies} at 59.0\%, and \texttt{extends} weakest (52.4\%)---reflecting that ``extends'' relations are often implicit and harder to validate from surface features alone. Negative-control analysis confirms clear separability from random non-edges ($p{=}0.004$). A matched-control analysis further shows that \texttt{extends} and \texttt{improves} are above matched baselines, while \texttt{applies} is near baseline---indicating that the pipeline is better at detecting methodological refinement than cross-domain transfer.

\noindent\textbf{Evidence-gated retrieval.} Converting protocol labels into a retrieval gate provides a controlled non-degradation diagnostic: the gate removes approximately 103 noisy semantic candidates per topic while changing mean recall from 0.164 to 0.182 and precision from 0.105 to 0.117. Per topic, recall improves only on \texttt{Memory in the Age of AI Agents} ($+0.071$) and is exactly unchanged on the other three topics at stored precision; precision improves on two topics and exactly ties on two, with no declines. Removing candidates can improve top-$5$ recall because unsupported semantic edges may place noisy papers in the highest-ranked slots; filtering them allows an external-proxy target to enter the fixed top-$5$ set. Given four topics, we interpret the mean gains descriptively rather than as statistical superiority. A target-role pruning heuristic further removes 10\% of semantically misclassified edges with no retrieval degradation. A ROC-style analysis over the 108 annotated edges (Figure~\ref{fig:roc_gate} in Appendix) yields AUC=0.75, with the optimal confidence threshold at 0.35 achieving F1=0.81. Per-type analysis reveals that \texttt{improves} and \texttt{applies} are well-separated (F1$>$0.88) while \texttt{extends} requires a higher threshold for acceptable precision.

\noindent\textbf{Human validation.} Two graduate-student annotators independently labeled the full 108-row stratified annotation packet (three relation types $\times$ three confidence tiers $\times$ 12 edges per cell) following detailed guidelines with worked examples. Each row includes source/target paper context, the inferred relation with confidence, and a matched non-edge control. Annotators assigned ternary labels: \textit{supported}, \textit{partial}, or \textit{unsupported}.

Inter-annotator agreement reaches Cohen's $\kappa{=}0.64$ (substantial) for the binary supported-or-partial versus unsupported decision; three-way agreement is lower ($\kappa{=}0.38$), so we use ternary labels primarily for failure-mode diagnosis rather than as a calibrated ordinal scale. After adjudication of 45 disagreement cases by a third annotator, the final consensus labels show 68.5\% supported-or-partial overall (fully supported: 40.7\%; partial: 27.8\%; unsupported: 31.5\%). Partial edges retain correct directionality and semantic relevance but lack complete lexical evidence---the evidence gate (Section~\ref{sec:mekg_construction}) is specifically designed to filter these into usable versus noisy subsets. By relation type: \texttt{improves} is strongest at 80.6\%, \texttt{applies} reaches 72.2\%, and \texttt{extends} is weakest at 52.8\%---consistent with the deterministic protocol's ranking. By confidence tier: high-confidence edges reach 75.0\% supported-or-partial, mid-confidence 72.2\%, and low-confidence 58.3\%, confirming that the pipeline's confidence score is a meaningful quality signal. We release the full annotation packet with adjudicated labels and guidelines.

\subsection{Citation Grounding Audits (RQ4)}

\noindent\textbf{Claim-level anchoring.} We sample 20 cited sentences per run from eight generated surveys (160 total), stratified across single- and multi-citation cases. Each sentence is compared against cited paper titles, abstracts, and cached text using lexical overlap measures. The audit finds 93 anchored, 36 partial, and 31 weak/missing cases (80.6\% anchored-or-partial).

To validate that lexical anchoring reflects genuine evidence alignment, we construct a hard-negative contrast: for each cited sentence, we compare similarity to the cited paper against a same-run non-cited paper matched on topic. Cited evidence outperforms hard negatives in 65.4\% of cases, with anchored sentences at 71.7\% and weak cases near chance (48.4\%). This gradient confirms that the lexical audit captures real grounding quality.

\noindent\textbf{Section-level alignment.} A 42-section audit across all eight runs shows zero uncited sections, with 85.7\% achieving aligned-or-partial status and mean citation density of 19.5 citations per section.

\subsection{Baseline Comparisons}

SurveyGraph produces substantially richer outlines than SciSage (7.25 vs.\ 2.75 sections on same topics), with SciSage systematically missing key structural roles such as methodology comparison and future directions. Structural comparisons against 48 released outputs from four concurrent systems are in Appendix~\ref{app:repro_audit}.

\noindent\textbf{SurveyBench-facing probes.} We do not claim full SurveyBench leaderboard parity. Instead, we report two benchmark-facing diagnostics that are fully replayable from released artifacts. First, a reference-list probe over the local SurveyForge SurveyBench snapshot gives the released SurveyForge references 0.465 mean coverage and human survey references 0.629, while topic-mismatched SurveyGraph archives overlap too weakly to support an end-to-end benchmark claim. Second, a live collection-stage probe with the repository's enhanced collector covers four SurveyBench topics under a fixed eight-paper budget and reaches 0.521 mean reference coverage versus 0.166 for exact-topic retrieval on the same topics. These probes show benchmark-facing collection viability, but they do not test outlining, long-form generation, or claim support. Full details appear in Appendix~\ref{app:surveybench_detail}. Separately, a same-backend pipeline-completion audit runs the full collection--MEKG--outline--generation--PDF path on five topics with Qwen-2.5 and a fixed 30-paper budget; all five complete, with 31--202 typed relations and 7--8 generated sections (Appendix~\ref{app:five_topic_completion}). This audit supports cross-topic pipeline execution, not SurveyBench leaderboard parity or long-form quality superiority.

\noindent\textbf{Comparison with concurrent systems.} We use MVSS \citep{liu2026mvss} and StructSurvey \citep{pedinotti2026structsurvey} as closest structure-first points of comparison, but we do not present the current comparison as a same-backend live rerun. The available evidence is bounded: released numbers, inspectable artifacts, and our own relation-diversity diagnostics indicate that SurveyGraph's distinctive contribution is the typed semantic layer and its evidence gate, not an uncontested end-to-end quality win. We therefore use these systems to contextualize the design space and to motivate typed diagnostics; full same-topic, same-budget, same-backend cross-system generation parity remains future work.

\subsection{Generation-Stage Controlled Diagnostics (RQ5)}

To separate backend confounds from retrieval-stage evidence within a bounded API budget, we ran a same-backend Qwen diagnostic batch on the topic \textit{Large Language Model Function Calling}. The full run built a 30-paper MEKG with 74 typed relations; the controlled diagnostics reused this substrate where possible and disabled external multi-judge evaluation, AI image generation, and optional critical-analysis LLM paragraph rewriting. Table~\ref{tab:qwen_generation_diagnostics} reports the artifact-backed component results. We also completed a matched flat generation run that fixes the same 30-paper corpus, Qwen backend, top-down outline, and generation/reflection settings while removing MEKG relations and graph-derived attributes.

\begin{table*}[t]
\centering
\scriptsize
\setlength{\tabcolsep}{3pt}
\begin{tabularx}{\textwidth}{>{\raggedright\arraybackslash}p{0.16\textwidth} c c c c c >{\raggedright\arraybackslash}X}
\toprule
\textbf{Variant} & \textbf{Sec./pg.} & \textbf{Tab./vis.} & \textbf{Rubric} & \textbf{MEKG} & \textbf{Global} & \textbf{Boundary condition} \\
\midrule
Full Qwen & 7/17 & 4/4 & 4.135 & .562 & 3.80 & Complete core pipeline artifact; external evaluator was stopped after PDF generation and is not claimed. \\
No outline critic / no post-refine & 7/18 & 4/5 & 4.110 & .619 & 3.80 & Bounded by the globally disabled text post-refiner; isolates outline-critic removal under that setting. \\
No evolutionary analyzer & 7/17 & 4/-- & 4.135 & .562 & 3.50 & Post-visual diagnostic with evolutionary analyzer disabled, not a full content regeneration. \\
No hybrid outline & 5/23 & 4/4 & 3.955 & .779 & 3.20 & Same 30-paper/74-relation substrate; bounded LLM outline timeout fell back to rule-based MEKG outline. \\
\bottomrule
\end{tabularx}
\caption{Same-backend generation-stage controlled diagnostics on one Qwen topic. Values are reported for the archived diagnostic batch described in Appendix~\ref{app:generation_diagnostics}. \textit{MEKG} is the internal graph-native overall score; \textit{Global} is the global-reflection score. These diagnostics support bounded component analysis rather than a leaderboard claim or cross-system superiority statement.}
\label{tab:qwen_generation_diagnostics}
\end{table*}

The diagnostic pattern is informative but deliberately not overinterpreted. The full run and no-critic/no-refine run both remain in the ``good'' rubric band, suggesting that the MEKG substrate and reflective content generator provide much of the artifact quality under this topic. Disabling the evolutionary analyzer mainly lowers the global-reflection score (3.80 to 3.50), indicating that the evolutionary narrative contributes to consistency checks more than to the internal graph metrics in this controlled setting. The no-hybrid outline run completes successfully but shifts to a simpler five-section organization and lower rubric/global-reflection scores. This directly supports the design choice to keep hybrid outline generation as the default while exposing a reproducible fallback path when the LLM outline call fails.

\begin{table}[t]
\centering
\scriptsize
\setlength{\tabcolsep}{2pt}
\begin{tabular}{@{}lccc@{}}
\toprule
\textbf{Check} & \textbf{SurveyGraph} & \textbf{Flat} & \textbf{$\Delta$} \\
\midrule
Rubric score & 4.135 & 4.030 & +0.105 \\
Citation grounding & 89.1\% (49/55) & 86.3\% (44/51) & +2.8 pp \\
Section alignment & 75.0\% (3/4) & 60.0\% (3/5) & directional \\
Global reflection & 3.80 & 3.20 & +0.60 \\
\bottomrule
\end{tabular}
\caption{Matched generation diagnostic on one Qwen topic. Both conditions use the same 30 papers, backend, outline, and generation/reflection settings; the flat condition removes MEKG relations and graph-derived attributes. Grounding and alignment use the complete enhanced survey bodies.}
\label{tab:matched_flat_generation}
\end{table}

The matched result improves the local rubric, global reflection, and citation grounding. Section alignment is 3/4 versus 3/5: the numerator is identical and the eligible-section denominator differs by one, so this row is treated only as directional evidence. Neither condition has an uncited eligible section. The evidence-level TF--IDF hard-negative contrast instead favors flat generation (65.5\% vs. 78.4\%). We therefore do not claim universal metric dominance. The controlled evidence supports a narrower mechanism: MEKG improves structural evidence organization and grounding, while generic lexical similarity can favor the flatter output.

\subsection{Discussion}

The seven-method benchmark reveals a complementarity pattern: typed MEKG retrieval does not dominate graph-free methods on surface recall, but provides typed relations that explain retrieval decisions, enable evidence-based filtering, and support safer fusion that boosts structural density without sacrificing recall. The CoCit metric confirms these gains reflect genuine bibliometric relationships rather than artifacts of the graph's own edge definitions.

\begin{figure}[t]
    \centering
    \includegraphics[width=\columnwidth]{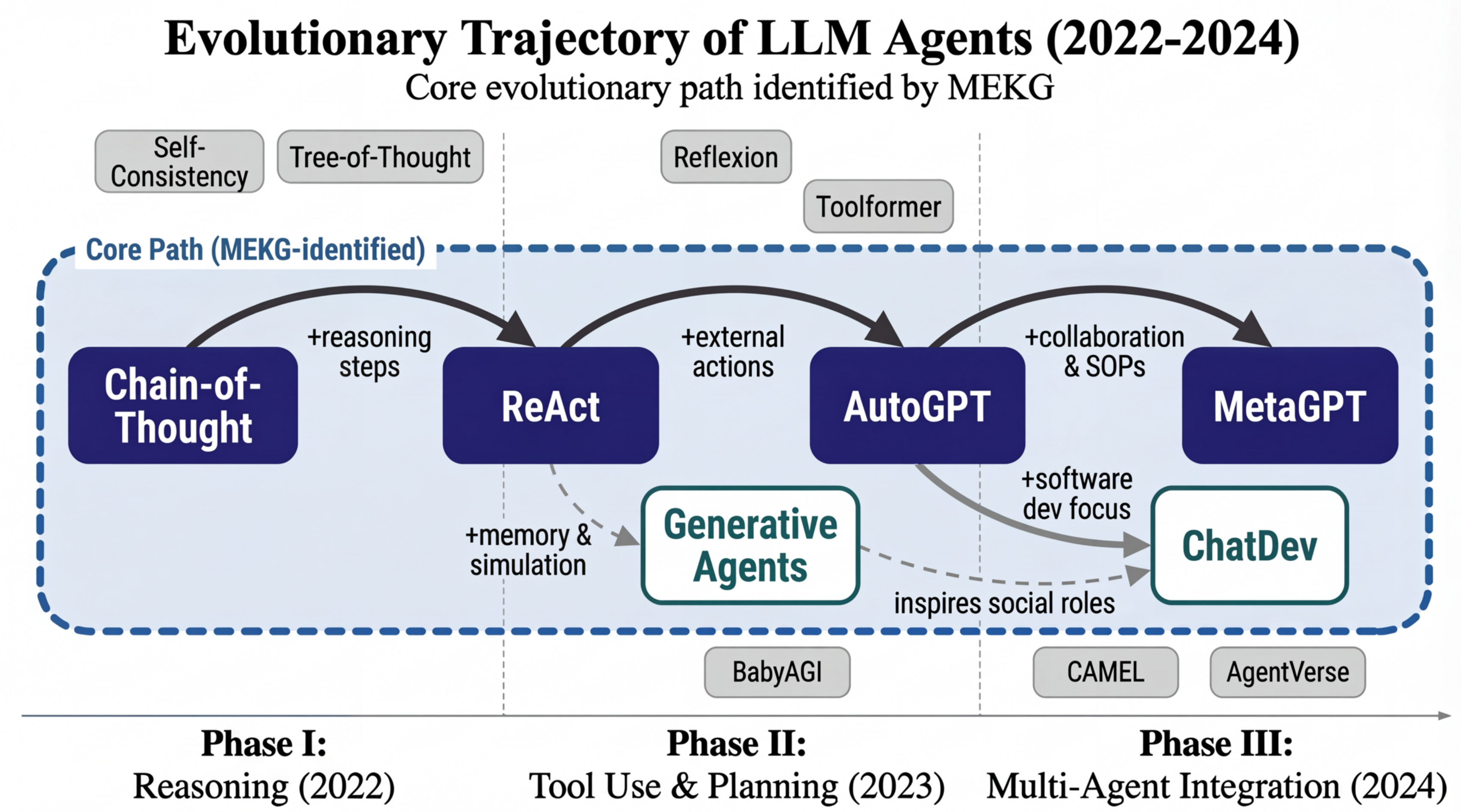}
    \caption{MEKG-driven evolutionary structure: the graph captures method-level follow-on, competition, and shared-resource relations, enabling the narrative synthesizer to trace research lineages.}
    \label{fig:evolution}
\end{figure}

Beyond retrieval, MEKG structure directly informs outline generation---graph-aware outlines average 7.25 sections with 36.5 nodes vs.\ 2.75 sections from SciSage on the same topics. The SurveyBench-facing collection probes show that the collector can recover benchmark references, while the retrieval and signal-component ablations isolate where the strongest evidence currently lies. In particular, removing 1-hop expansion collapses recall to near-flat levels, while removing PageRank centrality has a smaller effect, suggesting that MEKG's value lies primarily in its connectivity structure rather than in node-level features alone. The same-backend Qwen diagnostics in Tables~\ref{tab:qwen_generation_diagnostics} and~\ref{tab:matched_flat_generation} add generation-stage checks: hybrid outlines and evolutionary analysis improve artifact organization/consistency signals, but the evidence remains a bounded one-topic diagnostic rather than a broad benchmark.

The concurrent-system comparison is used for positioning rather than end-to-end superiority. SurveyGraph's distinguishing capability is relation-specific filtering: its typed semantic layer exposes separate evidence thresholds for \texttt{extends}, \texttt{improves}, and \texttt{applies}. The ROC analysis (Appendix~\ref{app:roc_gate}) quantifies this precision--recall control, while MVSS and StructSurvey provide complementary structure-first designs under different evaluation settings.

\noindent\textbf{Scalability and cost considerations.} MEKG construction accounts for 55\% of the pipeline's token budget, but this cost is amortized across multiple retrieval queries, outline variants, and generation passes. Once built, a topic MEKG serves repeated survey generation at marginal cost (retrieval + synthesis only). The modular architecture supports drop-in replacement of the LLM backbone, enabling cost--quality tradeoffs without pipeline redesign. In practice, once a MEKG is constructed for a topic, it can serve multiple survey variants and incremental updates as new papers appear.

\section{Conclusion}\label{sec:conclusion}

We have presented SurveyGraph, a graph-centric framework for automated survey generation that uses a Methods-Entity Knowledge Graph to model method-level evolutionary structure. Our evaluation demonstrates three key findings: (1) graph-neighborhood traversal drives the gain over flat retrieval on the archived graph-native core diagnostic ($p{=}0.004$); (2) MEKG-anchored fusion provides a strong recall--structure tradeoff in a controlled seven-method benchmark, supported by an independent co-citation metric; and (3) evidence gating (AUC=0.75) filters noisy semantic candidates, with human annotation ($\kappa{=}0.64$ for binary support) confirming that 68.5\% of sampled edges are supported-or-partial after adjudication. SurveyBench results remain collection-stage diagnostics, and MVSS/StructSurvey remain contextual comparisons. The supported contribution is an auditable typed structural layer for safer fusion, relation-specific filtering, and controllable generation.

These findings also define how the framework should be used. The graph is an auditable control layer: typed relations expose structural assumptions behind retrieval, and relation-specific gates prevent unsupported candidates from reaching synthesis. External co-citation and citation-grounding checks provide independent signals, while the same-backend study isolates component behavior without making a bounded diagnostic a leaderboard claim.

\section*{Limitations}
\label{sec:limitations}

Our SurveyBench-facing results are collection-stage and reference-list probes rather than full end-to-end benchmark submissions (Section~\ref{sec:experiments}). They cover four enhanced-collector topics and all ten reference-list topics, but they do not yet include full outlining, writing, or claim-support evaluation on the complete SurveyBench suite, nor evaluations on SurGE or SurveyLens. We plan to complete these evaluations using the same infrastructure released with this paper.

The semantic-edge validation covers 108 stratified edges with two annotators and adjudication. While the inter-annotator agreement on the binary support decision ($\kappa{=}0.64$) is substantial, the sample covers three relation types from a single domain cluster and the three-way label agreement is lower. The four-topic SurveyBench collection probe spans distinct AI subfields, and CoCit improvements are consistent across the archived retrieval runs, but these are indirect generalization checks rather than substitutes for a larger human edge study. Extending the annotation to additional topics and relation types would further strengthen these claims. The released annotation package is designed to enable this community effort.

Our comparison with MVSS and StructSurvey is not a same-backend live rerun. Differences in model backbones, released artifacts, and evaluation settings introduce a confound, so we do not use these comparisons as decisive evidence of end-to-end superiority. The same-backend Qwen diagnostics in Section~\ref{sec:experiments} reduce confounding for SurveyGraph's own generation-stage choices, but they do not replace a same-topic, same-budget, same-backend rerun against MVSS or StructSurvey. The robust architectural claim is narrower: SurveyGraph exposes typed relation diversity and per-relation evidence gates that flat or citation-tree systems do not provide; full controlled cross-system generation parity remains future work.

Finally, the graph-native metrics (WCR, CC, SC) are designed as diagnostic signals for understanding pipeline behavior rather than as standalone quality measures. We complement them with external-facing metrics (citation grounding, section alignment, evidence contrasts) that do not depend on the internal graph.

\section*{Ethical Considerations}
\label{sec:ethical_considerations}

SurveyGraph operates on publicly available scholarly metadata, paper text, and generated survey artifacts; we do not report experiments involving human subjects or newly collected personal data. We use only standard bibliographic information already exposed by scholarly records, and we do not release any additional person-level annotations beyond the derived experimental artifacts discussed in this paper. The main ethical risk is that automated survey generation may amplify hallucinated citations, unsupported summaries, or structurally plausible but incomplete synthesis if the graph or evaluator is incomplete. For this reason, we explicitly separate diagnostic graph metrics from benchmark-facing evaluation and avoid presenting the internal composite score as a proxy for human scholarly judgment. We also used AI assistants during code exploration and drafting support, but all reported claims were checked against archived artifacts, logs, and reproduced outputs before inclusion.

\section*{Acknowledgments}

This work is supported by the Intelligent Computing Center of Shenzhen University.

\label{page:references}
\bibliography{custom}

\clearpage
\appendix
\label{page:appendix}
\renewcommand{\thetable}{A.\arabic{table}}
\setcounter{table}{0}
\renewcommand{\thefigure}{A.\arabic{figure}}
\setcounter{figure}{0}
\input{appendix}

\end{document}

%% file: appendix.tex
% ==========================================
% APPENDIX - SurveyGraph Supplementary Material
% ==========================================

\section{Artifact Inventory}
\label{app:artifacts}

This appendix summarizes the SurveyGraph evaluation artifacts. All numerical results reported in the main text are backed by archived outputs and reproducible scripts released with the supplementary material. Code and artifacts are available at: \url{https://github.com/xzzvsxd/SurveyGraph}.

\paragraph{Reproducibility package.}
The released package includes: (i) eight completed survey generation runs with full MEKG graphs, evaluator traces, and rendered outputs; (ii) retrieval benchmark replay scripts for all seven methods; (iii) the filled 108-row semantic-edge annotation file and guidelines; (iv) citation-grounding, evidence-contrast, section-alignment, and evidence-gate audits with cached results; and (v) baseline comparison artifacts. The inventory below names the exact files and cached outputs associated with the reported analyses so that each result can be traced to its corresponding artifact.

\begin{table*}[t]
\centering
\small
\setlength{\tabcolsep}{4pt}
\begin{tabular}{lrrrrrr}
\toprule
\textbf{Archived run} & \textbf{Overall} & \textbf{Ref.} & \textbf{Struct.} & \textbf{Content} & \textbf{Ethics} & \textbf{Media} \\
\midrule
Large Language Model Function Calling & 76.77 & 77.50 & 77.40 & 64.83 & 82.17 & 89.24 \\
Document Visual Question Answering & 75.19 & 78.58 & 60.77 & 72.50 & 82.77 & 85.65 \\
Memory in the Age of AI Agents & 80.94 & 81.15 & 64.67 & 79.23 & 95.50 & 90.57 \\
Embodied Question Answering in City Airspace (run 1) & 77.17 & 87.44 & 57.50 & 74.94 & 80.37 & 86.81 \\
Text-to-SQL Agent & 80.65 & 94.00 & 61.96 & 75.07 & 80.75 & 92.55 \\
Embodied Question Answering in City Airspace (run 2) & 71.60 & 82.30 & 47.85 & 66.82 & 74.94 & 90.10 \\
Approach of Document Question Answer & 73.50 & 78.98 & 59.06 & 65.25 & 81.50 & 89.39 \\
Agent & 78.09 & 100.00 & 56.33 & 66.72 & 81.25 & 86.41 \\
\midrule
\textbf{Mean} & \textbf{76.74} & \textbf{84.99} & \textbf{60.69} & \textbf{70.67} & \textbf{82.41} & \textbf{88.84} \\
\bottomrule
\end{tabular}
\caption{Dimension-level scores for the eight completed local SurveyGraph artifacts. Scores come from the repository's current internal evaluator and are therefore useful for run-to-run comparison under the same implementation, not as public-benchmark leaderboard numbers.}
\label{tab:artifact_dimensions_app}
\end{table*}

\begin{table*}[t]
\centering
\small
\setlength{\tabcolsep}{5pt}
\begin{tabular}{lrrrrr}
\toprule
\textbf{Archived run} & \textbf{WCR} & \textbf{CC} & \textbf{SC} & \textbf{TC} & \textbf{Bib. Int.} \\
\midrule
Large Language Model Function Calling & 0.25 & 1.00 & 0.50 & 0.75 & 1.00 \\
Document Visual Question Answering & 0.42 & 0.83 & 0.17 & 1.00 & 1.00 \\
Memory in the Age of AI Agents & 0.37 & 1.00 & 0.17 & 0.83 & 1.00 \\
Embodied Question Answering in City Airspace (run 1) & 0.70 & 1.00 & 0.20 & 1.00 & 0.86 \\
Text-to-SQL Agent & 0.80 & 1.00 & 0.20 & 1.00 & 1.00 \\
Embodied Question Answering in City Airspace (run 2) & 0.41 & 1.00 & 0.00 & 1.00 & 1.00 \\
Approach of Document Question Answer & 0.30 & 1.00 & 0.17 & 0.83 & 1.00 \\
Agent & 1.00 & 1.00 & 0.33 & 1.00 & 1.00 \\
\midrule
\textbf{Mean} & \textbf{0.53} & \textbf{0.98} & \textbf{0.22} & \textbf{0.93} & \textbf{0.98} \\
\bottomrule
\end{tabular}
\caption{Graph-native diagnostics and closely related reference metrics for the same eight artifacts. The low mean SC relative to CC illustrates why structure metrics should currently be interpreted as brittle diagnostics rather than as final quality proof.}
\label{tab:artifact_diagnostics_app}
\end{table*}

\section{Reproducibility Audit}
\label{app:repro_audit}

\subsection{Offline Evaluation Rerun}

We replayed the evaluator on an archived function-calling artifact using the repository's fast configuration and with LLM-based judging disabled. The rerun produced an overall score of 78.14, with reference quality 77.50, structural soundness 70.50, content integrity 79.00, ethics/integrity 97.92, and rich media quality 68.17. This confirms that the evaluator is replayable on archived outputs, but it also shows that aggregate scores depend on whether LLM-assisted dimensions are active.

\subsection{Fresh Uploaded-PDF End-to-End Smoke Runs}

A pair of fresh smoke runs was executed with the same single uploaded PDF under low-cost GLM and Qwen settings. In both runs, external search and most optional refinement stages were disabled to minimize cost while preserving end-to-end completion. The system successfully:
\begin{itemize}[nosep,leftmargin=*]
    \item parsed one uploaded PDF,
    \item completed the full generation pipeline,
    \item rendered a final survey PDF together with \LaTeX{}/Bib artifacts.
\end{itemize}
These runs are intentionally minimal and therefore generate one-paper graphs with zero inter-paper relations, so they are not evidence of graph richness. Their role is narrower: to verify that fresh uploaded-PDF ingestion, outline generation, section synthesis, and final rendering all complete in the present repository under low-cost settings.

\begin{table}[t]
\centering
\small
\setlength{\tabcolsep}{4pt}
\begin{tabularx}{\columnwidth}{>{\raggedright\arraybackslash}X c c c c}
\toprule
\textbf{Fresh run} & \textbf{Pages} & \textbf{Sections} & \textbf{Relations} & \textbf{Fast overall} \\
\midrule
GLM single-PDF smoke & 40 & 7 & 0 & 79.28 \\
Qwen single-PDF smoke & 36 & 7 & 0 & 74.30 \\
\bottomrule
\end{tabularx}
\caption{Fresh low-cost end-to-end SurveyGraph smoke reruns. These are one-paper completions designed to verify pipeline completion rather than benchmark-level graph quality.}
\label{tab:fresh_smoke_app}
\end{table}

\subsection{Same-Backend Generation-Stage Controlled Diagnostics}
\label{app:generation_diagnostics}

To reduce model/backend confounding in generation-stage ablations, we ran a cost-capped Qwen diagnostic batch on \textit{Large Language Model Function Calling} and retained its run manifest with the archived results. The batch uses \texttt{max\_parallel=1}, disables the external multi-judge evaluator, disables AI image generation, and keeps optional critical-analysis LLM paragraph enhancement off. The goal is bounded component evidence, not a SurveyBench leaderboard result or a same-backend rerun against MVSS/StructSurvey.

\begin{table*}[t]
\centering
\scriptsize
\setlength{\tabcolsep}{3pt}
\begin{tabularx}{\textwidth}{>{\raggedright\arraybackslash}p{0.15\textwidth} c c c c c >{\raggedright\arraybackslash}p{0.24\textwidth} >{\raggedright\arraybackslash}X}
\toprule
\textbf{Run} & \textbf{Papers/rel.} & \textbf{Sec./pages} & \textbf{Tables/vis.} & \textbf{Rubric} & \textbf{MEKG} & \textbf{Artifact summary} & \textbf{Scope boundary} \\
\midrule
Full Qwen & 30/74 & 7/17 & 4/4 & 4.135 good & .5617 & \path{qwen_full_artifact_summary.json} & Core pipeline completed; external evaluator was stopped after PDF generation, so no external-evaluator score is claimed. \\
No outline critic / no post-refine & 30/74 & 7/18 & 4/5 & 4.110 good & .6186 & \path{qwen_no_critic_ablation_summary.json} & Bounded by globally disabled text post-refinement; isolates outline-critic removal under that setting. \\
No evolutionary analyzer & 30/74 & 7/17 & 4/-- & 4.135 good & .5617 & \path{qwen_no_evolutionary_ablation_summary.json} & Post-visual-stage diagnostic; disables evolutionary analysis rather than regenerating all content from scratch. \\
No hybrid outline & 30/74 & 5/23 & 4/4 & 3.955 acceptable & .7795 & \path{qwen_no_hybrid_ablation_summary.json} & Reuses the same MEKG; after bounded LLM outline timeout, falls back to a rule-based MEKG outline with five sections. \\
\bottomrule
\end{tabularx}
\caption{Artifact-backed same-backend generation-stage diagnostics. All rows are Qwen runs on the same topic with the same 30-paper/74-relation substrate where applicable. \textit{Rubric} is the local rubric evaluator score; \textit{MEKG} is the internal graph-native overall score. ``--'' means the summary did not record that field.}
\label{tab:generation_diagnostics_app}
\end{table*}

The most useful signal is organizational rather than a simple score ordering. The full and no-critic/no-refine variants both remain in the good rubric band, indicating that the shared MEKG substrate and reflective content generator carry much of the artifact quality. Disabling the evolutionary analyzer leaves the internal MEKG score unchanged in this post-visual diagnostic but lowers global reflection from 3.80 to 3.50. The no-hybrid run completes and even improves the internal MEKG diagnostic because it cites a different subset of core papers, but its global reflection falls to 3.20 and the output shifts to a simpler five-section structure. We therefore use the diagnostic to support the default hybrid-outline design, while explicitly treating the no-hybrid result as a bounded fallback/control rather than a pristine collection-stage rerun. The non-monotonic ordering between the internal MEKG score and the rubric/global-reflection outcomes is itself informative: graph-native coverage and final-artifact quality measure different properties. Accordingly, the MEKG score is used only as an internal diagnostic, relation density is not treated as a final quality metric, and quality-facing evidence comes from grounding, hard-negative contrast, section alignment, and controlled generation behavior.

For the matched flat diagnostic in Table~\ref{tab:matched_flat_generation}, the flat run reuses the same 30-paper Function Calling corpus, Qwen configuration, top-down outline, and reflection settings, but its serialized graph contains zero relations and graph-derived paper attributes are stripped before generation. The final-body audit evaluates all recoverable cited sentences rather than a favorable subset. SurveyGraph improves the global-reflection score by 0.60 and grounding by 2.8 percentage points. Section alignment is 3/4 versus 3/5, with the same numerator and a one-section denominator difference, so it is retained only as a directional diagnostic. The hard-negative lexical contrast favors flat by 13.0 points. This mixed pattern is retained in the artifact and motivates the structural, rather than across-metric, interpretation in the main text.

\subsection{Five-Topic Same-Backend Pipeline Completion}
\label{app:five_topic_completion}

To separate pipeline execution from the collection-stage SurveyBench probes, we additionally ran five complete surveys with Qwen-2.5 under the same 30-paper budget. Each run completed collection, MEKG construction, outlining, long-form generation, and PDF production. These artifacts support cross-topic pipeline completion only; they are not matched SurveyBench leaderboard runs and are not used to claim long-form quality superiority.

\begin{table}[t]
\centering
\small
\setlength{\tabcolsep}{3pt}
\begin{tabularx}{\columnwidth}{>{\raggedright\arraybackslash}X r r r}
\toprule
\textbf{Topic} & \textbf{Papers} & \textbf{Relations} & \textbf{Sections} \\
\midrule
LLM-based Multi-Agent & 30 & 121 & 7 \\
Evaluation of LLMs & 30 & 31 & 7 \\
RAG for LLMs & 30 & 104 & 7 \\
Multimodal LLMs & 30 & 190 & 8 \\
Instruction Tuning & 30 & 202 & 7 \\
\bottomrule
\end{tabularx}
\caption{Five-topic same-backend pipeline-completion audit. All rows completed PDF production; the table reports structural scale rather than a quality score.}
\label{tab:five_topic_completion_app}
\end{table}

\subsection{Cross-Framework Artifact-Level Comparison}

To strengthen the baseline side without over-claiming fair rerun parity, we additionally analyzed released artifacts bundled inside the local ARISE repository snapshot. This yielded 48 baseline artifacts that are directly inspectable without new generation cost: 3 AutoSurvey PDFs, 10 SurveyForge PDFs, 8 SurveyX PDFs, and 27 ARISE outputs. We then compared them with the two fresh SurveyGraph reruns at the artifact level, using lightweight structural and completeness signals such as page count, extracted text length, citation-marker count, section-heading count, figure/table mentions, and whether nearby \LaTeX{}/Bib/review artifacts were available. We treat this comparison as an \emph{external-anchor artifact audit}: it contextualizes the scale and completeness of SurveyGraph outputs against released artifacts from public systems, but it is not a same-topic, same-budget, same-backend live benchmark.

\begin{table*}[t]
\centering
\small
\setlength{\tabcolsep}{5pt}
\begin{tabularx}{\textwidth}{>{\raggedright\arraybackslash}p{0.19\textwidth} r r r r r >{\raggedright\arraybackslash}X}
\toprule
\textbf{System} & \textbf{Count} & \textbf{Pages mean} & \textbf{Text chars mean} & \textbf{Citation mean} & \textbf{Section mean} & \textbf{Completeness signal} \\
\midrule
SurveyGraph-GLM-fresh & 1 & 40.00 & 169009 & 94.00 & 31.00 & TeX/Bib present \\
SurveyGraph-Qwen-fresh & 1 & 36.00 & 149616 & 19.00 & 31.00 & TeX/Bib present \\
AutoSurvey & 3 & 31.00 & 89238 & 221.33 & 0.33 & PDF-only released artifacts \\
SurveyForge & 10 & 27.80 & 186486 & 406.50 & 39.50 & PDF-only released artifacts \\
SurveyX & 8 & 23.00 & 86917 & 273.12 & 32.88 & PDF-only released artifacts \\
ARISE & 27 & 29.78 & 204664 & 453.30 & 60.19 & 100\% TeX/Bib/review; 96.3\% multi-round \\
\bottomrule
\end{tabularx}
\caption{System-level summary of the released artifact comparison. This table should be read as an artifact-backed structural/completeness audit, not as a fair live rerun benchmark.}
\label{tab:artifact_level_compare_app}
\end{table*}

The main value of this comparison is twofold. First, it shows that SurveyGraph fresh reruns produce complete inspectable writing artifacts rather than only intermediate logs. Second, it provides a low-cost external anchor for artifact scale and completeness against public systems whose released outputs are already bundled in the workspace. It does \emph{not} justify quality-ranking claims across systems because the evidence mixes fresh reruns and released outputs from heterogeneous topics.

\subsubsection{Released-Artifact Dedup Robustness Check}

Because released artifact folders may contain copied PDFs, we ran one extra SHA-based dedup check over the 50 total artifact-comparison rows (48 released artifacts plus the two fresh SurveyGraph reruns). This found 4 exact-PDF duplicate groups (4 removable rows total), affecting ARISE, SurveyForge, and SurveyX. After deduplication, system-level means moved only modestly---for example, ARISE citation mentions shifted from 453.30 to 436.84, SurveyForge from 406.50 to 424.89, and SurveyX from 273.12 to 264.57---without reversing the qualitative comparison story. We therefore keep the original artifact table for transparency while treating the cross-framework comparison as structurally informative rather than numerically precise.

\begin{table}[t]
\centering
\small
\setlength{\tabcolsep}{4pt}
\begin{tabularx}{\columnwidth}{>{\raggedright\arraybackslash}X c c c c c}
\toprule
\textbf{System} & \textbf{Orig.} & \textbf{Dedup} & \textbf{Page $\Delta$} & \textbf{Cit. $\Delta$} & \textbf{Sec. $\Delta$} \\
\midrule
ARISE & 27 & 25 & -0.18 & -16.46 & -0.31 \\
SurveyForge & 10 & 9 & +0.64 & +18.39 & +0.50 \\
SurveyX & 8 & 7 & -0.43 & -8.55 & -0.73 \\
\bottomrule
\end{tabularx}
\caption{Released-artifact dedup robustness check. Exact-PDF duplicates were removed by SHA-256 prefix; the resulting shifts are modest and do not change the paper's qualitative artifact-level comparison claims.}
\label{tab:artifact_dedup_app}
\end{table}

\subsubsection{Structural-Proxy Robustness Check}

Because the artifact-level comparison relies on cheap PDF-structure proxies, we ran one more zero-cost audit over the same 50 rows. The goal is not to claim that heading extraction is perfect, but to test whether the comparison is dominated by a brittle parser. Across all 50 rows, 47 exposed non-zero table-of-contents (TOC) entries. Among those 47 rows, the extracted TOC count matched the extracted section-heading count exactly in all cases (100.0\%). The only heading-count mismatch in the full table came from one AutoSurvey PDF that had no recoverable TOC but one detected heading. References headings were recoverable in 94.0\% of all rows and in every artifact from ARISE, SurveyForge, SurveyX, and the two fresh SurveyGraph reruns. This supports a narrow interpretation: the artifact-level structural comparison still depends on proxy metrics, but those proxies are internally consistent for nearly all released/fresh PDFs that expose a usable document structure.

\begin{table}[t]
\centering
\small
\setlength{\tabcolsep}{3pt}
\begin{tabularx}{\columnwidth}{>{\raggedright\arraybackslash}p{0.37\columnwidth} c c c c}
\toprule
\textbf{System} & \textbf{N} & \textbf{TOC} & \textbf{TOC=Sec.} & \textbf{Ref.} \\
\midrule
ARISE & 27 & 27 & 100.0\% & 100.0\% \\
AutoSurvey & 3 & 0 & 0.0\% & 0.0\% \\
SurveyForge & 10 & 10 & 100.0\% & 100.0\% \\
SurveyGraph-GLM-fresh & 1 & 1 & 100.0\% & 100.0\% \\
SurveyGraph-Qwen-fresh & 1 & 1 & 100.0\% & 100.0\% \\
SurveyX & 8 & 8 & 100.0\% & 100.0\% \\
\bottomrule
\end{tabularx}
\caption{Structural-proxy robustness audit over the 50 artifact-level comparison rows. Cheap structure proxies are internally consistent for nearly all artifacts that expose a usable TOC signal.}
\label{tab:artifact_structure_proxy_app}
\end{table}

\subsubsection{Length-Normalized Artifact Proxy Check}

To test whether the artifact-level comparison is driven only by raw PDF length, we ran one more zero-cost audit over the same comparison table after SHA-based deduplication. This leaves 46 unique rows. For each system, we normalize citation-marker counts by words (per 1k words), section headings by words (per 10k words), and figure+table mentions by words (per 10k words), while also tracking whether editable \LaTeX{}/Bib artifacts are available. The result is intentionally narrow: it does not rank quality, but it does check whether the previously reported structural differences survive a simple document-length normalization.

The normalized pattern still differs across systems. ARISE remains structurally dense after normalization, with a median heading density of 22.10 per 10k words and a figure/table density of 9.69 per 10k words, while also preserving a 100\% editable-artifact rate. SurveyGraph's two fresh reruns remain much shorter in citation density (2.56 per 1k words combined), but they still retain non-zero heading density (14.81 per 10k words), non-zero figure/table density (13.12 per 10k words), and a 100\% editable-artifact rate. AutoSurvey remains near zero on normalized heading and figure/table density, whereas SurveyX stays high on both normalized citation and heading density. This suggests that the cross-framework artifact comparison is not only a page-length effect.

\begin{table}[t]
\centering
\scriptsize
\setlength{\tabcolsep}{2pt}
\begin{tabularx}{\columnwidth}{>{\raggedright\arraybackslash}p{0.31\columnwidth} c c c c c}
\toprule
\textbf{System} & \textbf{N} & \textbf{Cit./1k} & \textbf{Head./10k} & \textbf{Fig+Tbl./10k} & \textbf{Editable} \\
\midrule
ARISE & 25 & 14.95 & 22.10 & 9.69 & 100.0\% \\
AutoSurvey & 3 & 16.68 & 0.00 & 0.00 & 0.0\% \\
SurveyForge & 9 & 14.75 & 14.28 & 0.00 & 0.0\% \\
SurveyGraph-fresh & 2 & 2.56 & 14.81 & 13.12 & 100.0\% \\
SurveyX & 7 & 22.28 & 28.14 & 13.45 & 0.0\% \\
\bottomrule
\end{tabularx}
\caption{Length-normalized artifact proxy audit over the 46 deduplicated comparison rows. The released/fresh structural comparison is not explained only by raw document length.}
\label{tab:artifact_length_normalized_app}
\end{table}

\subsection{SciSage Outline-Only Live Partial Baseline}

We executed four live SciSage partial baselines by reusing its native outline-generation path only, patching the model backend to a DashScope-compatible endpoint, and skipping full section writing, RAG retrieval, and full-paper synthesis. The final matched set contains four \emph{same-topic} low-cost runs on \texttt{Document Visual Question Answering}, \texttt{Text-to-SQL Agent}, \texttt{Large Language Model Function Calling}, and \texttt{Memory in the Age of AI Agents} with the DashScope-backed \texttt{qwen3.6-plus} chat model. These experiments remain outline-stage partial baselines, but they are materially stronger than the earlier topic-adjacent check because both topic and backend family are matched to archived SurveyGraph slices on four topics.

\begin{table}[t]
\centering
\scriptsize
\setlength{\tabcolsep}{1pt}
\begin{tabular}{>{\raggedright\arraybackslash}p{0.15\columnwidth} >{\raggedright\arraybackslash}p{0.18\columnwidth} >{\raggedright\arraybackslash}p{0.11\columnwidth} >{\raggedright\arraybackslash}p{0.16\columnwidth} >{\raggedright\arraybackslash}p{0.21\columnwidth}}
\toprule
\textbf{System} & \textbf{Setup} & \textbf{Ext-core} & \textbf{Outline} & \textbf{Runtime / note} \\
\midrule
SGraph & archived DocVQA & 0.083 @20 & 7 sec / 34 n. & 133.4 min; 24 refs \\
SciSage & live DocVQA outline-only & -- & 3 sec / 9 n. & 117.2 s; gate passed \\
\bottomrule
\end{tabular}
\caption{One real same-backend matched subset on \emph{Document Visual Question Answering}. This is not full-paper parity: SurveyGraph is an archived full run, while SciSage is a live outline-only run.}
\label{tab:scisage_matched_subset_app}
\end{table}

We also ran one extra zero-cost structure-only comparison over the same matched DocVQA pair. Instead of scoring content quality, this check aligns the top-level section titles to a coarse set of survey roles (introduction/scope, task background, motivation, methods, evaluation, challenges, conclusion). Under this title-role view, the archived SurveyGraph outline covers 7 unique roles, while the live SciSage outline covers 4. The longest shared ordered role subsequence has length 4 (\textit{intro/scope} $\rightarrow$ \textit{task/background} $\rightarrow$ \textit{methods} $\rightarrow$ \textit{challenges}), which yields a 0.571 SciSage role-recall against the SurveyGraph outline template for this topic. The missing roles are \textit{motivation}, \textit{evaluation}, and \textit{conclusion}. We treat this as a coarse structural-coverage probe only: it does not score subsection quality, retrieval evidence, or full-paper usefulness.

\begin{table}[t]
\centering
\scriptsize
\setlength{\tabcolsep}{2pt}
\begin{tabular}{>{\raggedright\arraybackslash}p{0.18\columnwidth} >{\raggedright\arraybackslash}p{0.16\columnwidth} >{\raggedright\arraybackslash}p{0.16\columnwidth} >{\raggedright\arraybackslash}p{0.15\columnwidth} >{\raggedright\arraybackslash}p{0.26\columnwidth}}
\toprule
\textbf{Topic} & \textbf{SGraph} & \textbf{SciSage} & \textbf{Role recall} & \textbf{Missing roles} \\
\midrule
DocVQA & 7 sec / 34 nodes & 3 sec / 9 nodes & 0.571 & motivation, evaluation, conclusion \\
Text-to-SQL & 7 sec / 40 nodes & 3 sec / 9 nodes & 0.429 & task background, motivation, other, conclusion \\
\midrule
LLM function calling & 8 sec / 34 nodes & 3 sec / 9 nodes & 0.429 & task background, motivation, other, conclusion \\
Memory in AI agents & 7 sec / 38 nodes & 2 sec / 4 nodes & 0.286 & task background, motivation, other, challenges, conclusion \\
\midrule
Mean & 7.25 sec / 36.5 nodes & 2.75 sec / 7.75 nodes & 0.429 & motivation and conclusion recur across all 4 \\
\bottomrule
\end{tabular}
\caption{Four same-backend-family matched subsets aggregated across \emph{Document Visual Question Answering}, \emph{Text-to-SQL Agent}, \emph{Large Language Model Function Calling}, and \emph{Memory in the Age of AI Agents}. Three of the four live SciSage runs pass their reflection gate; the memory-topic run fails and is also the structurally narrowest.}
\label{tab:scisage_two_topic_summary_app}
\end{table}

The second matched topic, \texttt{Text-to-SQL Agent}, follows the same qualitative pattern while adding topic diversity at very low cost. The archived SurveyGraph slice exposes a 5-paper external-core proxy target set with 0.400 top-20 proxy recall together with a 7-section/40-node outline, whereas the live SciSage outline-only run returns a 3-section/9-node outline in 132.27 seconds and passes the same reflection gate. We then add matched runs on \texttt{Large Language Model Function Calling} and \texttt{Memory in the Age of AI Agents}. The memory-topic baseline is useful precisely because it is a negative case: the archived SurveyGraph slice still exposes a 7-section/38-node outline, while the live SciSage run collapses to only 2 top-level sections and 4 total nodes in 140.17 seconds and fails SciSage's own reflection gate. Aggregated over the four matched topics, the live SciSage runs have mean latency 131.97 seconds, mean role-recall 0.429, and a 0.75 reflection-gate pass rate, but still leave motivation- and conclusion-like roles systematically under-covered relative to the archived SurveyGraph outlines. We again stress that this is a structure-stage matched-subset comparison only, not a full-paper quality benchmark.

To add one more benchmark-facing check at near-zero cost, we first replayed the archived section-to-paper retrieval diagnostic on two matched topics and then expanded that replay to a four-topic replay that better matches the paper's structural claim. The stronger replay compares three retrieval settings on the same archived topic slices and survey-reference external proxy targets: flat lexical retrieval, a simpler structural/meta graph that removes evolutionary semantic edges, and the full typed MEKG. On \texttt{Document Visual Question Answering}, recall rises from 0.125 (flat) to 0.125 (simple graph) to 0.167 (full graph), with precision rising from 0.130 to 0.136 to 0.190. On \texttt{Large Language Model Function Calling}, both graph variants recover one proxy hit where flat retrieval recovers none (0.000$\rightarrow$0.077). On \texttt{Memory in the Age of AI Agents} and \texttt{Text-to-SQL Agent}, recall remains tied, but both graph variants still increase relation density relative to flat retrieval. Averaged across the four matched topics, flat retrieval reaches 0.135 mean top-$5$ external-proxy recall and 0.087 precision, the simpler graph reaches 0.154 recall and 0.090 precision, and the full typed graph reaches 0.164 recall and 0.105 precision. Relation density rises from 0.120 to 0.163 and 0.165 respectively. Because the sample is only four topics, the paired tests are still inconclusive, but the replay addresses the simpler-baseline question with controlled evidence.

\begin{table}[t]
\centering
\small
\setlength{\tabcolsep}{2pt}
\begin{tabularx}{\columnwidth}{>{\raggedright\arraybackslash}X c c c}
\toprule
\textbf{Topic / metric} & \textbf{Flat@5} & \textbf{Simple@5} & \textbf{Full@5} \\
\midrule
LLM function calling recall & 0.000 & 0.077 & 0.077 \\
DocVQA ext-proxy recall & 0.125 & 0.125 & 0.167 \\
Memory in AI agents recall & 0.214 & 0.214 & 0.214 \\
Text-to-SQL ext-proxy recall & 0.200 & 0.200 & 0.200 \\
\midrule
Mean ext-proxy recall & 0.135 & 0.154 & 0.164 \\
Mean ext-proxy precision & 0.087 & 0.090 & 0.105 \\
Mean relation density & 0.120 & 0.163 & 0.165 \\
\bottomrule
\end{tabularx}
\caption{Four-topic matched replay of the archived retrieval diagnostic against survey-reference external proxy targets. The simpler structural/meta graph already improves relation density over flat retrieval, while the full typed graph adds a small further gain in both mean recall and mean precision.}
\label{tab:matched_external_proxy_replay_app}
\end{table}

We also added a seven-run GraphRAG-lite matched baseline to avoid comparing typed MEKG retrieval only against flat lexical retrieval. This baseline builds a local TF-IDF k-nearest-neighbor graph from each run's paper titles, abstracts, and category strings, applies the same one-hop graph ranker and top-$5$ per-section budget, and evaluates against the same weak survey-reference external proxy. The result is intentionally mixed and therefore useful: the TF-IDF kNN graph obtains higher weak-proxy recall than both flat lexical retrieval and typed full-MEKG retrieval (0.347 vs.\ 0.252 and 0.153), while typed MEKG variants expose much denser neighborhoods under the archived MEKG relation yardstick. This confirms that simple similarity graphs are strong retrieval baselines on the current proxy and that the typed MEKG should be claimed as an auditable structural representation rather than as a universally better recall mechanism. The full output is saved under the GraphRAG-lite directory in the artifact map above.

\begin{table}[t]
\centering
\small
\setlength{\tabcolsep}{2pt}
\begin{tabularx}{\columnwidth}{>{\raggedright\arraybackslash}p{0.28\columnwidth} c c c c}
\toprule
\textbf{Method} & \textbf{R@5} & \textbf{P@5} & \textbf{MEKG dens.} & \textbf{Act. dens.} \\
\midrule
Flat lexical & 0.252 & 0.132 & 0.134 & 0.000 \\
TF-IDF kNN graph & 0.347 & 0.176 & 0.140 & 0.167 \\
Structural graph & 0.135 & 0.075 & 0.217 & 0.217 \\
Full typed MEKG & 0.153 & 0.089 & 0.228 & 0.228 \\
Safe fusion full & 0.261 & 0.134 & 0.179 & 0.179 \\
\bottomrule
\end{tabularx}
\caption{Seven-run GraphRAG-lite matched baseline replay. \emph{MEKG dens.} measures typed-MEKG connectivity among retrieved papers; \emph{Act. dens.} measures density in the graph used by the method. The TF-IDF kNN graph is a strong lightweight recall baseline on this weak proxy, while typed MEKG variants provide denser typed-neighborhood evidence.}
\label{tab:graphrag_lite_baseline_app}
\end{table}

To check whether the GraphRAG-lite baseline is a single-parameter accident, we additionally swept 16 TF-IDF kNN settings by varying the neighbor count $k\in\{2,3,5,8\}$ and minimum similarity threshold $\{0.00,0.05,0.10,0.15\}$. Every tested setting remains above the flat lexical recall baseline (0.252), with recall ranging from 0.274 to 0.347. The best and balanced setting is the one used above ($k=5$, threshold 0.05), with R@5 0.347 and P@5 0.176. This reinforces the conservative interpretation: simple similarity-graph retrieval is a strong baseline, so our typed-MEKG claim should emphasize structure, typed evidence, and auditability rather than raw weak-proxy recall dominance.

\begin{table}[t]
\centering
\small
\setlength{\tabcolsep}{3pt}
\begin{tabularx}{\columnwidth}{c c c c c}
\toprule
\textbf{k} & \textbf{Min sim.} & \textbf{R@5} & \textbf{P@5} & \textbf{Active rels} \\
\midrule
2 & 0.05 & 0.300 & 0.139 & 245.4 \\
3 & 0.05 & 0.291 & 0.134 & 358.6 \\
5 & 0.05 & 0.347 & 0.176 & 569.7 \\
8 & 0.05 & 0.342 & 0.165 & 859.1 \\
\midrule
Sweep min & -- & 0.274 & 0.133 & 27.0 \\
Sweep max & -- & 0.347 & 0.176 & 914.3 \\
\bottomrule
\end{tabularx}
\caption{GraphRAG-lite sensitivity sweep over 16 TF-IDF kNN settings. The table shows the threshold-0.05 slice plus the overall sweep range; all tested settings exceed the flat lexical recall baseline on the same weak external proxy.}
\label{tab:graphrag_lite_sweep_app}
\end{table}

\paragraph{Quality-aware and matched graph mini-benchmark.}
We also add an OpenAlex-backed bibliometric rerank baseline that uses title-matched \texttt{cited\_by\_count}, recency, and a lightweight paper-role prior, without using MEKG edges. Under the same seven-run weak survey-reference proxy and top-$5$ per-section budget, the strongest graph-free quality variant is lexical--bibliometric reciprocal-rank fusion (R@5/P@5 0.309/0.194), improving over flat lexical retrieval (0.252/0.132). This result is important because it shows that part of the retrieval gain can be obtained without typed relations; SurveyGraph's safer claim is therefore not raw recall superiority, but auditability and controllable fusion.

\begin{table}[t]
\centering
\small
\setlength{\tabcolsep}{2pt}
\begin{tabularx}{\columnwidth}{>{\raggedright\arraybackslash}p{0.32\columnwidth} c c c c}
\toprule
\textbf{Method} & \textbf{R@5} & \textbf{P@5} & \textbf{Quality} & \textbf{Cites} \\
\midrule
Flat lexical & 0.252 & 0.132 & 0.369 & 6.6 \\
OA quality only & 0.210 & 0.457 & 0.670 & 115.5 \\
Top-20 rerank & 0.280 & 0.156 & 0.393 & 11.7 \\
Flat + quality & 0.280 & 0.150 & 0.384 & 11.1 \\
Lexical--biblio RRF & 0.309 & 0.194 & 0.452 & 17.3 \\
\bottomrule
\end{tabularx}
\caption{OpenAlex-backed quality-aware retrieval baseline over the same seven-run weak survey-reference proxy. \emph{Quality} is the normalized citation/recency/role score among retrieved papers; \emph{Cites} is mean OpenAlex \texttt{cited\_by\_count}. This is a matched retrieval-stage baseline, not a full reproduction of quality-aware survey generation systems.}
\label{tab:bibliometric_rerank_app}
\end{table}

We then combined the strongest graph-free quality baseline with TF-IDF GraphRAG-lite and typed-MEKG retrieval in one equal-budget mini-benchmark. Table~\ref{tab:matched_quality_graph_benchmark_app} uses the same section queries, same top-$5$ budget, and same weak proxy for all methods. The best weak-proxy method is bibliometric-anchor + TF-IDF fusion (R@5/P@5 0.357/0.200). Full typed MEKG alone has lower weak-proxy recall but much higher typed-neighborhood density, while safe fusion recovers part of the recall without abandoning typed structure. This directly motivates our final positioning: simple quality/similarity baselines are strong, and typed MEKG is most useful when used as an auditable structure and fusion signal.

\begin{table}[t]
\centering
\small
\setlength{\tabcolsep}{1.8pt}
\begin{tabularx}{\columnwidth}{>{\raggedright\arraybackslash}p{0.30\columnwidth} c c c c}
\toprule
\textbf{Method} & \textbf{R@5} & \textbf{P@5} & \textbf{Align} & \textbf{Dens.} \\
\midrule
Flat lexical & 0.252 & 0.132 & 0.558 & 0.134 \\
Bibliometric RRF & 0.309 & 0.194 & 0.530 & 0.132 \\
TF-IDF GraphRAG-lite & 0.347 & 0.176 & 0.485 & 0.140 \\
SurveyGraph MEKG & 0.153 & 0.089 & 0.392 & 0.228 \\
Safe fusion MEKG & 0.261 & 0.134 & 0.528 & 0.179 \\
Biblio + TF-IDF fusion & 0.357 & 0.200 & 0.498 & 0.134 \\
Biblio + MEKG fusion & 0.304 & 0.178 & 0.486 & 0.174 \\
\bottomrule
\end{tabularx}
\caption{Seven-run matched quality/graph mini-benchmark. All methods use the same archived section queries and top-$5$ retrieval budget. \emph{Align} is a lexical section-evidence alignment proxy, and \emph{Dens.} is typed-MEKG relation density among retrieved papers.}
\label{tab:matched_quality_graph_benchmark_app}
\end{table}

\subsection{Baseline Audit Summary}

\begin{table}[t]
\centering
\small
\setlength{\tabcolsep}{4pt}
\begin{tabularx}{\columnwidth}{l l X}
\toprule
\textbf{System} & \textbf{Status} & \textbf{Decisive interpretation} \\
\midrule
SurveyForge & Partial & Released artifacts analyzable; live rerun still blocked by missing local \texttt{database/} and embedding assets \\
SurveyX & Analyzable & Released artifacts available, but no same-environment live rerun claimed \\
AutoSurvey & Analyzable & Released artifacts available, but no same-environment live rerun claimed \\
SciSage & Partial & Same-topic outline-only live baseline succeeds; full-paper parity still unverified \\
ARISE & Partial & Released artifacts analyzable; code path still expects extra provider/runtime setup for live rerun \\
\bottomrule
\end{tabularx}
\caption{Updated baseline audit summary after adding released artifact analysis and four same-topic live partial baselines. None of these systems should yet be counted as fully and fairly reproduced under a unified environment.}
\label{tab:baseline_audit_app}
\end{table}

Together, these checks justify a narrow claim set: SurveyGraph has real archived outputs, fresh low-cost end-to-end completions, same-backend internal generation diagnostics, and a partially replayable comparison story that now includes released cross-framework artifacts plus four same-topic live partial baselines. Full same-topic, same-budget, same-backend cross-system baseline parity remains future work.

\subsection{SurveyBench Reference-List Probe}

We also run a zero-API probe against the local SurveyBench snapshot bundled with the SurveyForge repository. We reused the public benchmark script's citation-coverage definition: after date filtering, the score is the fraction of generated reference IDs that appear in the topic's benchmark reference set. This is still not an end-to-end SurveyGraph benchmark run, because our archived SurveyGraph topics are not the ten SurveyBench topics and the SurveyForge local code path requires missing database/embedding assets for a live matched rerun. It does, however, calibrate what the benchmark-format reference metric looks like on the released SurveyForge and human reference lists, and it quantifies how little of the present SurveyGraph archive can be reused as a proxy.

\begin{table}[t]
\centering
\small
\setlength{\tabcolsep}{3pt}
\begin{tabularx}{\columnwidth}{>{\raggedright\arraybackslash}X c c c}
\toprule
\textbf{Source} & \textbf{N} & \textbf{Cov.} & \textbf{Match} \\
\midrule
SurveyForge released refs & 10 & 0.465 & 50.0 \\
Human survey refs & 10 & 0.629 & 148.8 \\
SurveyGraph archive overlaps & 15 & 0.078 & 3.9 \\
\bottomrule
\end{tabularx}
\caption{SurveyBench reference-list probe using the local SurveyForge SurveyBench snapshot. SurveyForge and human rows are true benchmark-format reference-list evaluations. The SurveyGraph rows are only top-overlap checks between existing archived corpora and SurveyBench topics, not end-to-end benchmark runs.}
\label{tab:surveybench_reference_probe_app}
\end{table}

The released SurveyForge references obtain nontrivial but topic-variable coverage: 0.857 on \textit{3D Gaussian Splatting}, 0.598 on \textit{Generative Diffusion Models}, 0.435 on \textit{Retrieval-Augmented Generation for Large Language Models}, and 0.296 on \textit{LLM-based Multi-Agent}. By contrast, the strongest SurveyGraph archive-to-benchmark overlaps are 11 references for \textit{LLM-based Multi-Agent} from the function-calling archive and 10 references for \textit{Evaluation of Large Language Models}; these are far too small to support a benchmark claim. We include the result as a reproducible calibration and as a guardrail: future benchmark evaluation must run SurveyGraph on the SurveyBench topics with matched retrieval budgets and reference extraction, rather than treating unrelated archives as benchmark evidence.

To add one small amount of genuinely live benchmark-facing evidence without paying for new long-form generation, we also ran an exact-topic arXiv retrieval probe through the repository's own retriever over all 10 SurveyBench topics, capping each topic at 50 returned papers and then scoring the retrieved arXiv IDs with the same date-filtered overlap rule. This collection-stage probe is deliberately narrower than an end-to-end survey benchmark: it does not test outlining, writing, or claim support. It nevertheless shows that the repository retriever can recover nontrivial benchmark references from live search alone, reaching 0.245 mean overlap coverage and 5.1 matched references per topic. The strongest topics are \textit{Hallucination in Large Language Models} (0.500; 6/12 valid targets matched), \textit{3D Gaussian Splatting} (0.412; 7/17), and \textit{Multimodal Large Language Models} (0.357; 5/14), while \textit{Graph Neural Networks} remains low (0.051; 2/39) and \textit{LLM-based Multi-Agent} still yields 0/3 matched targets under the benchmark's date filter. We therefore use this probe only to demonstrate live benchmark-facing retrieval viability, not end-to-end benchmark parity.

To further strengthen benchmark-facing evidence without paying for full survey generation, we ran a second live collection-stage probe using the repository's \texttt{EnhancedPaperCollector} rather than exact-topic search alone. This probe covers four SurveyBench topics (\textit{LLM-based Multi-Agent}, \textit{Evaluation of Large Language Models}, \textit{Retrieval-Augmented Generation for Large Language Models}, and \textit{Multimodal Large Language Models}) under a fixed 8-paper collection budget. Mean benchmark-reference coverage reaches 0.521, compared with 0.166 for the exact-topic probe on the same four topics, and all four topics improve. The largest gain is on \textit{Retrieval-Augmented Generation for Large Language Models} (0.188$\rightarrow$1.000), while \textit{LLM-based Multi-Agent} rises from 0.000 to 0.333. Because this probe still measures only collection-stage overlap, we treat it as stronger retrieval evidence rather than as end-to-end benchmark success.

\begin{table}[t]
\centering
\small
\setlength{\tabcolsep}{3pt}
\begin{tabularx}{\columnwidth}{>{\raggedright\arraybackslash}X c c c c}
\toprule
\textbf{Topic} & \textbf{Budget} & \textbf{Enhanced} & \textbf{Exact-topic} & \textbf{$\Delta$} \\
\midrule
LLM-based Multi-Agent & 8 & 0.333 & 0.000 & +0.333 \\
Evaluation of LLMs & 8 & 0.250 & 0.120 & +0.130 \\
RAG for LLMs & 8 & 1.000 & 0.188 & +0.812 \\
Multimodal LLMs & 8 & 0.500 & 0.357 & +0.143 \\
\midrule
Mean & 8 & 0.521 & 0.166 & +0.355 \\
\bottomrule
\end{tabularx}
\caption{Live four-topic SurveyBench-facing collection-stage probe using the repository \texttt{EnhancedPaperCollector}. Even under a small 8-paper budget, the enhanced collector outperforms exact-topic retrieval on all four matched topics.}
\label{tab:enhanced_collector_benchmark_probe_app}
\end{table}

\section{Additional Low-Cost Diagnostics}
\label{app:diagnostic_additions}

To strengthen the evidence without additional long-form generation, we ran fourteen extra post-hoc analyses over the already saved artifacts: the original ten local diagnostics, plus an encoder-sensitivity replay, a graph-safe fusion sweep, a semantic-gate error analysis, and a local cost-stage replay. All fourteen analyses are fully local and reproducible from the current workspace, except for already completed live collection probes explicitly marked as live in Appendix~\ref{app:repro_audit}.

\subsection{Offline Retrieval-Side Graph Diagnostic}

We constructed a fixed-budget diagnostic ablation over the eight archived runs. For each non-introduction/non-conclusion top-level section, we retrieved the top-$5$ papers with (i) a flat lexical section-to-paper ranker and (ii) a graph-aware ranker that adds category/subcategory overlap, PageRank centrality, and 1-hop relation expansion from the strongest lexical seeds. We then compared the union of retrieved papers under graph-native targets: core-paper recall, community coverage, average PageRank exposure, and relation density among retrieved papers. This experiment is intentionally diagnostic and should not be read as a public benchmark.

\begin{table}[t]
\centering
\small
\setlength{\tabcolsep}{2pt}
\begin{tabularx}{\columnwidth}{>{\raggedright\arraybackslash}X c c c c}
\toprule
\textbf{Metric} & \textbf{Flat} & \textbf{\shortstack{Graph\\aware}} & \textbf{$\Delta$} & \textbf{\shortstack{Graph\\better}} \\
\midrule
Core recall & 0.216 & 0.373 & +0.157 & 8/8 \\
Comm. cov. & 0.758 & 0.863 & +0.104 & 5/8 \\
Avg PageRank & 0.481 & 0.630 & +0.149 & 8/8 \\
Relation dens. & 0.162 & 0.259 & +0.098 & 8/8 \\
Unique papers & 21.38 & 21.50 & +0.13 & 4/8 \\
\bottomrule
\end{tabularx}
\caption{Offline retrieval-side diagnostic ablation on the eight archived artifacts. Under the same top-$5$ section budget, the graph-aware ranker retrieves more central and more mutually connected paper sets on average.}
\label{tab:graph_diag_app}
\end{table}

An exact paired sign-flip check over the same eight archived run pairs shows that the retrieval-side gain is not carried by only one or two outliers. Core-paper recall, average PageRank exposure, and relation density are all positive in 8/8 non-zero pairs, each with one-sided $p=0.0039$ and two-sided $p=0.0078$. Community coverage is directionally positive but weaker: 5 positive pairs, 1 negative pair, and 2 ties, giving one-sided $p=0.0781$ and two-sided $p=0.1563$. We report these values as compact diagnostic evidence only, not as benchmark-grade significance claims.

We also ran one small signal-component ablation on the same archived retrieval setup to separate the contributions of category/subcategory overlap, PageRank centrality, and 1-hop relation expansion. The most important result is that 1-hop expansion carries most of the observed retrieval-side lift: removing expansion drops mean core recall from 0.373 to 0.219 and relation density from 0.259 to 0.161, nearly matching the flat lexical baseline (0.216/0.162). By contrast, removing centrality only weakens mean core recall to 0.364 and relation density to 0.254, while removing category/subcategory overlap is nearly neutral on the current archived runs. We therefore treat the present graph-aware retrieval gain as primarily an expansion effect, with centrality as a secondary preference signal and overlap as a weak tie-breaker under this archive slice.

\begin{table}[t]
\centering
\small
\setlength{\tabcolsep}{4pt}
\begin{tabularx}{\columnwidth}{>{\raggedright\arraybackslash}X c c c}
\toprule
\textbf{Variant} & \textbf{Core rec.} & \textbf{Avg PR} & \textbf{Rel. dens.} \\
\midrule
Full graph-aware & 0.373 & 0.630 & 0.259 \\
No overlap & 0.373 & 0.629 & 0.259 \\
No centrality & 0.364 & 0.621 & 0.254 \\
No expansion & 0.219 & 0.486 & 0.161 \\
Flat lexical & 0.216 & 0.481 & 0.162 \\
\bottomrule
\end{tabularx}
\caption{Signal-component ablation inside the archived retrieval diagnostic. Removing 1-hop expansion nearly collapses the graph-aware gains back to the flat lexical baseline, while removing overlap is almost neutral on the current archive slice.}
\label{tab:signal_component_ablation_app}
\end{table}

\subsection{Retrieval Budget Sweep}

To check whether the graph-aware gain depends on a single retrieval setting, we replayed the same diagnostic at three small budgets: top-3, top-5, and top-7 papers per section. The main pattern is that graph-aware retrieval stays ahead of flat lexical retrieval on core-paper recall and relation density at all three budgets, so the top-5 choice in the main text is best understood as a conservative middle point rather than a cherry-picked setting.

\begin{table}[t]
\centering
\small
\setlength{\tabcolsep}{4pt}
\begin{tabularx}{\columnwidth}{>{\raggedright\arraybackslash}X c c c c}
\toprule
\textbf{Budget} & \textbf{Flat rec.} & \textbf{Graph rec.} & \textbf{$\Delta$} & \textbf{Wins} \\
\midrule
top-3 & 0.113 & 0.288 & +0.176 & 7/8 \\
top-5 & 0.216 & 0.373 & +0.157 & 8/8 \\
top-7 & 0.288 & 0.505 & +0.217 & 8/8 \\
\bottomrule
\end{tabularx}
\caption{Retrieval budget sweep over the eight archived runs. The graph-aware advantage is stable across modest top-k budgets, so the top-5 choice in the main text is a conservative middle point rather than a cherry-picked setting.}
\label{tab:budget_sweep_app}
\end{table}

\subsubsection{Confidence-Threshold Robustness Sweep}

To test whether the graph-aware gain depends mainly on low-confidence edges, we replayed the same top-$5$ retrieval diagnostic after pruning the archived MEKG to only relations with confidence at least 0.5, 0.7, or 0.9. The gain weakens mildly under stricter pruning, but it remains positive on both core-paper recall and relation density even at the highest threshold. This makes the current retrieval result more robust than a purely weak-edge explanation would suggest.

\begin{table}[t]
\centering
\small
\setlength{\tabcolsep}{4pt}
\begin{tabularx}{\columnwidth}{>{\raggedright\arraybackslash}X c c c c}
\toprule
\textbf{Thr.} & \textbf{Keep rel.} & \textbf{Graph rec.} & \textbf{Rec. $\Delta$} & \textbf{Graph dens.} \\
\midrule
$\geq$0.5 & 99.0\% & 0.373 & +0.157 & 0.259 \\
$\geq$0.7 & 80.7\% & 0.336 & +0.120 & 0.248 \\
$\geq$0.9 & 39.1\% & 0.342 & +0.126 & 0.245 \\
\bottomrule
\end{tabularx}
\caption{Confidence-threshold robustness sweep over the eight archived runs under the same top-$5$ section budget. Even after keeping only higher-confidence relations, graph-aware retrieval remains stronger than the flat baseline on the main retrieval-side diagnostics.}
\label{tab:confidence_sweep_app}
\end{table}

\subsubsection{Depth Sweep: 1-Hop vs.\ 2-Hop}

To test whether a deeper traversal helps, we replayed the same retrieval diagnostic with 1-hop and 2-hop expansion. The 2-hop variant does retrieve denser neighborhoods, but it also increases search breadth substantially, lowers unique-paper yield, and slightly weakens community coverage on average. This supports keeping 1-hop as the default while treating 2-hop as an exploratory mode.

\begin{table}[t]
\centering
\small
\setlength{\tabcolsep}{4pt}
\begin{tabularx}{\columnwidth}{>{\raggedright\arraybackslash}X c c c}
\toprule
\textbf{Metric} & \textbf{Flat} & \textbf{1-hop} & \textbf{2-hop} \\
\midrule
Core-paper recall & 0.216 & 0.373 & 0.388 \\
Community coverage & 0.758 & 0.863 & 0.817 \\
Average PageRank & 0.481 & 0.630 & 0.659 \\
Relation density & 0.162 & 0.259 & 0.309 \\
Unique papers & 21.38 & 21.50 & 17.00 \\
Expanded nodes & -- & 32.02 & 73.78 \\
Traversed edges & -- & 243.55 & 2358.56 \\
\bottomrule
\end{tabularx}
\caption{1-hop vs.\ 2-hop depth sweep over the same eight archived runs. 2-hop expansion adds coverage inside the graph, but it also pulls in many more nodes and edges and reduces the number of unique papers retained under the fixed retrieval budget.}
\label{tab:depth_sweep_app}
\end{table}

\subsubsection{Relation-Family Ablation}

To isolate which edge families matter most under the same top-$5$ section budget, we replayed the 1-hop graph-aware ranker with four relation inventories: the full graph, a \texttt{no-support} variant that removes only generic \texttt{supports} edges, a \texttt{structural-only} variant that keeps support/meta relations but removes evolutionary semantic edges, and a \texttt{semantic-only} variant that keeps only \texttt{extends}/\texttt{improves}/\texttt{applies}/\texttt{refutes}. The result is mixed but informative: dropping generic support edges changes retrieval only slightly, whereas restricting the graph to semantic edges alone degrades both core recall and relation density. This supports the interpretation that the current retrieval gain comes from the combination of structural and semantic links rather than one relation family in isolation.

\begin{table}[t]
\centering
\small
\setlength{\tabcolsep}{4pt}
\begin{tabularx}{\columnwidth}{>{\raggedright\arraybackslash}X c c c c}
\toprule
\textbf{Variant} & \textbf{Core rec.} & \textbf{Comm. cov.} & \textbf{Avg PR} & \textbf{Rel. dens.} \\
\midrule
Full graph & 0.373 & 0.863 & 0.630 & 0.259 \\
No support & 0.389 & 0.912 & 0.633 & 0.246 \\
Struct.-only & 0.389 & 0.863 & 0.621 & 0.256 \\
Sem.-only & 0.291 & 0.829 & 0.545 & 0.115 \\
\bottomrule
\end{tabularx}
\caption{Relation-family retrieval ablation over the eight archived runs. Removing only generic \texttt{supports} edges changes the averages little, while a semantic-only graph loses both core-paper recall and relation density.}
\label{tab:relation_family_ablation_app}
\end{table}

\subsection{Core-Score Weight Sensitivity Audit}

Because the default core-paper score uses fixed hand-tuned weights, we first ran a local sensitivity audit over the same eight archived MEKGs. We replayed the score with four moderate reweightings (PageRank-heavy, citation-heavy, recency-heavy, and degree-heavy) and found that the overlap with the default top-$10$ remains 9.0--9.75 papers on average across variants. To address the saved artifacts' missing citation signal, we then queried OpenAlex for 1,060 unique archived titles, accepted title matches with similarity at least 0.86 and compatible publication year, and recomputed the default score with OpenAlex \texttt{cited\_by\_count} as a citation proxy. The mean OpenAlex match rate is 92.7\%, all eight runs recover nonzero citation signal, and the enriched top-$10$ overlaps the stored-field top-$10$ by 6.00 papers on average (range 4--7), indicating that citations materially adjust but do not completely replace the structural/recency ranking.

\begin{table}[t]
\centering
\small
\setlength{\tabcolsep}{4pt}
\begin{tabularx}{\columnwidth}{>{\raggedright\arraybackslash}X c c c}
\toprule
\textbf{Variant vs.\ default top-$10$} & \textbf{Mean overlap} & \textbf{Median} & \textbf{Range} \\
\midrule
PageRank-heavy & 9.50 & 9.50 & 9--10 \\
Citation-heavy & 9.75 & 10.00 & 9--10 \\
Recency-heavy & 9.75 & 10.00 & 9--10 \\
Degree-heavy & 9.00 & 9.00 & 8--10 \\
\midrule
Default core-set size & 16.12 & -- & -- \\
Elbow-selected size & 17.38 & -- & -- \\
Archived citation fields populated & 0/8 & -- & -- \\
OpenAlex non-zero signal & 8/8 & -- & -- \\
OpenAlex top-10 overlap & 6.00 & 6.50 & 4--7 \\
\bottomrule
\end{tabularx}
\caption{Archived core-score sensitivity and OpenAlex citation-proxy audit. Moderate weight perturbations preserve most of the default top-$10$ core papers, while OpenAlex title matching recovers a nonzero citation proxy for all eight runs and shows that bibliometric enrichment changes, but does not erase, the archived structural ranking.}
\label{tab:core_weight_sensitivity_app}
\end{table}

We then replayed the survey-reference external-core proxy under two stronger OpenAlex-backed settings. Keeping the original score weights but replacing dormant local citation fields with OpenAlex \texttt{cited\_by\_count} raises mean top-$20$ proxy recall from 0.162 to 0.189. A second leave-one-run-out coarse-grid calibration over the same proxy reaches 0.289 mean top-$20$ recall, typically favoring citation-heavy weights (most often 0.7--0.8 on the citation term). The gain is real but not yet decisive: the paired sign test is still weak, and \textit{Text-to-SQL Agent} collapses under the citation-only optimum selected on the remaining six runs. We therefore treat this replay as evidence that the present score form can be externally calibrated, not as proof that the current scorer is solved.

\begin{table}[t]
\centering
\footnotesize
\setlength{\tabcolsep}{3pt}
\begin{tabularx}{\columnwidth}{>{\raggedright\arraybackslash}p{0.28\columnwidth} c c >{\raggedright\arraybackslash}X}
\toprule
\textbf{Calibration} & \textbf{Mean} & \textbf{Median} & \textbf{Note} \\
\midrule
Archived default score & 0.162 & 0.077 & local citation field dormant \\
OpenAlex default weights & 0.189 & 0.125 & same score form, citation proxy repaired \\
LOO tuned & 0.289 & 0.308 & coarse PR/cit/rec/deg grid \\
\bottomrule
\end{tabularx}
\caption{OpenAlex-backed external-core proxy calibration over the seven archived runs with usable survey-reference targets. Replacing dormant local citation fields with OpenAlex \texttt{cited\_by\_count} gives a small default gain, while leave-one-run-out coarse-grid tuning yields a larger but still explicitly proxy-bound improvement.}
\label{tab:external_core_calibration_app}
\end{table}

\subsection{Archived Input / Title-Integrity Audit}

Because graph size or relation counts could potentially be inflated by trivial upstream duplication, we ran an additional audit directly over the archived MEKG paper/relation stores. This check is intentionally narrow: it does not claim full scholarly canonicalization, but it asks whether the saved runs exhibit obvious failure modes that should have been caught before any semantic interpretation.

Across the eight archived runs, the saved MEKG inputs contain 1,146 paper records and 18,160 relation records. Using the same lightweight title normalization rule implemented in the repository's online fetcher (lowercasing, punctuation stripping, and whitespace collapsing before title-hash deduplication), we found zero within-run exact title duplicates and zero within-run normalized-title duplicates. We also found zero missing values for the core paper fields we rely on in this paper (\texttt{id}, \texttt{title}, \texttt{abstract}, \texttt{category}, \texttt{subcategory}, \texttt{year}, \texttt{source}), zero missing \texttt{source}/\texttt{target}/\texttt{type}/\texttt{confidence} fields in the saved relation records, and zero dangling relation endpoints whose source or target paper ID is absent from the corresponding archived paper store. Relation confidence values in this archive slice range from 0.4 to 1.0, with a global mean of 0.806.

To clarify why some smaller runs still expose more than 2,300 relations, we also grouped the archived artifacts by corpus size and summarized relation density. The key pattern is that the 50-paper runs are denser in \emph{relations per paper}, not because of duplicate paper rows, but because compact corpora accumulate many repeated structural/meta links. In particular, the three high-frequency non-semantic types \texttt{supports}, \texttt{same\_author}, and \texttt{shares\_architecture} already account for roughly 69\% of edges in both small and large groups, while semantic edges remain a minority.

\begin{table}[t]
\centering
\small
\setlength{\tabcolsep}{4pt}
\begin{tabularx}{\columnwidth}{>{\raggedright\arraybackslash}X c c c c}
\toprule
\textbf{Run group} & \textbf{Papers} & \textbf{Relations} & \textbf{Rel./paper} & \textbf{Sem. share} \\
\midrule
50-paper runs & 50.0 & 2314.67 & 46.29 & 10.1\% \\
$\sim$200-paper runs & 199.2 & 2243.20 & 11.26 & 12.4\% \\
\midrule
\multicolumn{5}{p{0.95\columnwidth}}{\footnotesize In both groups, \texttt{supports} + \texttt{same\_author} + \texttt{shares\_architecture} account for about 69\% of all edges, and structural/meta relation families exceed 87\% overall.} \\
\bottomrule
\end{tabularx}
\caption{Grouped relation-density summary for the archived MEKG artifacts. Smaller 50-paper runs are denser in raw relations per paper, but the density is dominated by repeated structural/meta links rather than by a larger semantic-edge share.}
\label{tab:relation_density_grouped_app}
\end{table}

This audit should be interpreted conservatively. It does \emph{not} show that all synonymy or cross-run aliasing problems are solved, and it does not validate the semantic correctness of the edges themselves. What it does show is that, within each archived run used in this paper, the graph is not being trivially enlarged by exact duplicate rows, normalized-title duplicates, or broken relation references.

\subsection{Semantic-Edge Consistency Audit}

We further audited the semantic edge types most central to MEKG's evolutionary claims: \texttt{extends}, \texttt{improves}, and \texttt{applies}. For each archived edge, we checked whether the source paper is temporally at least as recent as the target paper, whether the source title/abstract contains an explicit lexical cue compatible with the relation type, and what confidence score was saved in the archived relation record. This remains a consistency audit rather than a human-annotated precision/recall study.

\begin{table}[t]
\centering
\small
\setlength{\tabcolsep}{2pt}
\begin{tabularx}{\columnwidth}{l r c c c}
\toprule
\textbf{Relation} & \textbf{Count} & \textbf{Temp.} & \textbf{Cue} & \textbf{Conf.} \\
\midrule
\texttt{extends} & 1276 & 99.8\% & 24.4\% & 0.673 \\
\texttt{improves} & 366 & 99.7\% & 65.8\% & 0.675 \\
\texttt{applies} & 449 & 98.0\% & 24.7\% & 0.620 \\
\bottomrule
\end{tabularx}
\caption{Archived semantic-edge consistency audit over 2,091 MEKG edges. High temporal-direction consistency supports the intended evolutionary orientation, while the lower cue-support rates indicate that many saved edges rely on contextual inference rather than only keyword matching.}
\label{tab:edge_audit_app}
\end{table}

\subsubsection{Semantic-Edge Negative-Control Audit}

The consistency checks above still do not answer whether the saved semantic edges are distinguishable from plausible non-edges. We therefore added a deterministic negative-control audit over the same 2,091 archived \texttt{extends}/\texttt{improves}/\texttt{applies} edges. For each observed edge, we compared source--target lexical overlap against two controls from the same archived run: a \emph{matched non-edge} with compatible temporal direction and target-role class, and an intentionally \emph{easy non-edge} chosen to have minimal lexical overlap with the source. This design uses no new model calls and is deliberately conservative: the matched control is meant to be hard, not a strawman.

\begin{table}[t]
\centering
\small
\setlength{\tabcolsep}{3pt}
\begin{tabularx}{\columnwidth}{>{\raggedright\arraybackslash}X c c c c}
\toprule
\textbf{Control} & \textbf{Edge} & \textbf{Control} & \textbf{$\Delta$} & \textbf{Better} \\
\midrule
Matched non-edge & 0.0748 & 0.0733 & +0.0015 & 1109/2091 \\
Easy non-edge & 0.0748 & 0.0249 & +0.0500 & 2079/2091 \\
\bottomrule
\end{tabularx}
\caption{Semantic-edge negative-control audit using lexical overlap. Observed semantic edges are clearly separated from arbitrary low-overlap non-edges, but they are nearly tied with temporally and target-role matched non-edge controls. This negative result supports our conservative interpretation that lexical evidence alone cannot validate \texttt{extends}/\texttt{improves}/\texttt{applies} edges.}
\label{tab:semantic_negative_control_app}
\end{table}

At the run level, the easy-control contrast is positive in 8/8 runs (one-sided sign-flip $p=0.0039$), while the matched-control contrast is positive in only 5/8 runs (one-sided $p=0.1719$). By relation type, \texttt{extends} and \texttt{improves} are slightly above matched controls on average, whereas \texttt{applies} is slightly below matched controls. We therefore do not use this audit to claim semantic precision. Its value is diagnostic: it rules out completely arbitrary low-overlap links, but confirms that a benchmark-grade edge study needs human labels, citation-context spans, and negative sampling beyond local lexical overlap.

\subsubsection{Confidence-Stratified Semantic-Edge Profile}

To test whether the archived confidence scores align with local textual evidence, we add one more zero-cost profile over the same 2,091 semantic edges. We pruned the archived semantic layer to edges with confidence at least 0.5, 0.7, or 0.9 and recomputed the same consistency statistics. The result is modest but still informative: higher-confidence subsets retain near-perfect temporal direction, and explicit source-cue support rises slightly from 32.0\% at $\geq 0.5$ to 33.2\% at $\geq 0.7$ before becoming sample-limited at $\geq 0.9$ (52 edges total). We therefore interpret the saved confidence field as a useful but still imperfect diagnostic priority signal rather than as calibrated probabilistic correctness.

\begin{table}[t]
\centering
\small
\setlength{\tabcolsep}{3pt}
\begin{tabularx}{\columnwidth}{>{\raggedright\arraybackslash}X c c c c}
\toprule
\textbf{Min conf.} & \textbf{Count} & \textbf{Temp.} & \textbf{Cue} & \textbf{Avg conf.} \\
\midrule
$\geq$0.5 & 2035 & 99.6\% & 32.0\% & 0.668 \\
$\geq$0.7 & 1078 & 99.7\% & 33.2\% & 0.754 \\
$\geq$0.9 & 52 & 100.0\% & 30.8\% & 0.907 \\
\bottomrule
\end{tabularx}
\caption{Confidence-stratified semantic-edge consistency profile over the archived MEKG. Higher-confidence subsets retain near-perfect temporal direction, while explicit lexical cue support rises only modestly, which supports treating confidence as a diagnostic ranking signal rather than as a calibrated correctness score.}
\label{tab:edge_conf_profile_app}
\end{table}

\subsubsection{Repeated-Protocol Semantic-Edge Audit}

To add an extra sanity check without running a full human study, we expanded the semantic-edge check into a stratified repeated-protocol audit with 72 rows stratified by relation type and confidence bucket (24 each for \texttt{extends}, \texttt{improves}, and \texttt{applies}; 24 each for low/mid/high confidence). Unlike the earlier 36-row author-side audit, this extension uses a deterministic protocol over archived artifacts: each sampled row is assigned a support tier from temporal direction, source-side lexical cue support, target-role admissibility, and whether the observed edge beats a temporally and role-matched non-edge control in lexical overlap. This remains a zero-API deterministic audit rather than a human precision/recall study, but it is stricter and more reproducible than the earlier spot check.

\begin{table}[t]
\centering
\small
\setlength{\tabcolsep}{3pt}
\begin{tabularx}{\columnwidth}{l c c c c}
\toprule
\textbf{Relation} & \textbf{Supported} & \textbf{Partial} & \textbf{Unsupported} & \textbf{Total} \\
\midrule
\texttt{extends} & 5 & 6 & 13 & 24 \\
\texttt{improves} & 3 & 14 & 7 & 24 \\
\texttt{applies} & 4 & 12 & 8 & 24 \\
\midrule
\textbf{Overall} & \textbf{12} & \textbf{32} & \textbf{28} & \textbf{72} \\
\bottomrule
\end{tabularx}
\caption{Stratified repeated-protocol audit over 72 stratified semantic-edge rows. The protocol is deterministic and stricter than the earlier author-side spot audit, so the lower support rate should be read as a more conservative view of semantic-edge reliability rather than as contradictory evidence.}
\label{tab:spot_audit_app}
\end{table}

In this stricter 72-row audit, the supported-or-partial rate is 61.1\% overall. By relation type, \texttt{improves} is the strongest at 70.8\%, \texttt{applies} reaches 66.7\%, and \texttt{extends} is weakest at 45.8\%. The dominant failure modes are missing explicit source cues (18 rows), weak lineage signal under the matched-control contrast (17), lexical overlap not better than the matched control despite otherwise admissible roles/time (14), and target-role mismatch (10). Only one sampled row is a temporal reversal. By confidence bucket, the low and high buckets each reach 66.7\% supported-or-partial, while the mid bucket is weaker at 54.2\%, reinforcing our interpretation that the archived confidence field is at best a rough priority signal rather than a calibrated correctness estimate.

To make the protocol less abstract, Table~\ref{tab:semantic_snippet_examples_app} shows representative passed, partial, and failed cases with snippets extracted from paper metadata fields. These examples are not additional human labels; they illustrate why we treat semantic edges as candidate evidence. Supported cases usually contain source-side method or adaptation cues and beat the matched non-edge control, while failed cases often point from a method paper to a benchmark/dataset target or have only generic topical overlap.

\begin{table*}[t]
\centering
\scriptsize
\setlength{\tabcolsep}{3pt}
\begin{tabularx}{\textwidth}{l l >{\raggedright\arraybackslash}p{0.20\textwidth} >{\raggedright\arraybackslash}p{0.20\textwidth} >{\raggedright\arraybackslash}p{0.18\textwidth} >{\raggedright\arraybackslash}X}
\toprule
\textbf{Rel.} & \textbf{Label} & \textbf{Source} & \textbf{Target} & \textbf{Failure / support mode} & \textbf{Local source snippet} \\
\midrule
\texttt{extends} & supported & Adapting While Learning: Grounding LLMs for Scientific Tool Use & Confucius: Iterative Tool Learning from Introspection Feedback by Easy-to-Difficult Curriculum & well-supported protocol case & ``Tool Usage Adaptation (TUA)'' categorizes problems by model accuracy and trains on adapted tool-use settings. \\
\texttt{extends} & unsupported & Adapting While Learning: Grounding LLMs for Scientific Tool Use & API-Bank: A Comprehensive Benchmark for Tool-Augmented LLMs & target-role mismatch & The source discusses tool-use adaptation, but the target is a benchmark resource rather than a precursor method. \\
\texttt{improves} & supported & LaTr: Layout-Aware Transformer for Scene-Text VQA & Spatially Aware Multimodal Transformers for TextVQA & well-supported protocol case & The source explicitly proposes a layout-aware multimodal architecture for scene-text VQA. \\
\texttt{improves} & unsupported & LayoutLMv2: Multi-modal Pre-training for Visually-Rich Document Understanding & DocVQA: A Dataset for VQA on Document Images & target-role mismatch & The source describes pre-training objectives, but the target is a dataset paper, so the relation is not a direct method refinement. \\
\texttt{applies} & supported & Advancing SLM Tool-Use Capability using Reinforcement Learning & MetaTool Benchmark for Large Language Models & well-supported protocol case & The source discusses reinforcement learning for tool-augmented agents in a setting tied to tool-use evaluation. \\
\texttt{applies} & partial & CheMatAgent: Enhancing LLMs for Chemistry and Materials & Toolformer: Language Models Can Teach Themselves to Use Tools & lexical-overlap not better than matched control & The source is a domain application of tool-use ideas, but the local evidence is not enough to establish a direct transfer path. \\
\bottomrule
\end{tabularx}
\caption{Representative semantic-edge examples from the snippet audit. The table illustrates both positive and negative protocol decisions and motivates the evidence-gated replay in Table~\ref{tab:semantic_evidence_gate_app}.}
\label{tab:semantic_snippet_examples_app}
\end{table*}

\subsubsection{Generated-Claim Citation-Support and Lexical Anchor Audits}

Because grounding ultimately depends on generated survey statements rather than only graph edges, we expanded the earlier two-topic spot check into a deterministic lexical-anchor audit over eight archived SurveyGraph outputs. The expanded audit samples 20 cited body sentences per run (160 sentences total), including both single- and multi-citation sentences, and compares each sentence with cited paper titles, abstracts, and cached text. The labels are \emph{anchored}, \emph{partial}, or \emph{weak/missing}; they are lexical grounding triage rather than blinded human citation-faithfulness annotation.

\begin{table}[t]
\centering
\small
\setlength{\tabcolsep}{2pt}
\begin{tabularx}{\columnwidth}{>{\raggedright\arraybackslash}p{0.33\columnwidth} c c c c c}
\toprule
\textbf{Topic} & \textbf{Cand.} & \textbf{Samp.} & \textbf{Anch.} & \textbf{Part.} & \textbf{Weak} \\
\midrule
LLM function calling & 242 & 20 & 11 & 7 & 2 \\
Doc. visual QA & 232 & 20 & 13 & 4 & 3 \\
Memory in AI agents & 288 & 20 & 14 & 2 & 4 \\
Embodied QA run 1 & 175 & 20 & 11 & 3 & 6 \\
Text-to-SQL agent & 202 & 20 & 5 & 8 & 7 \\
Embodied QA run 2 & 217 & 20 & 13 & 6 & 1 \\
Approach of DocQA & 215 & 20 & 12 & 4 & 4 \\
Agent & 138 & 20 & 14 & 2 & 4 \\
\midrule
All & 1709 & 160 & 93 & 36 & 31 \\
\bottomrule
\end{tabularx}
\caption{Expanded claim-to-citation lexical anchor audit over eight archived generated surveys. The overall anchored-or-partial rate is 129/160 = 80.6\%, with 103 single-citation and 57 multi-citation sentences. Weak/missing-anchor rows mark higher-risk claims for manual review rather than automatic hallucination labels.}
\label{tab:expanded_claim_anchor_audit_app}
\end{table}

For continuity with earlier checks, we also keep the smaller two-topic author-side spot audit: 24 single-citation body sentences from the archived DocQA and Text-to-SQL artifacts yielded 15 supported, 7 partial, and 2 unsupported cases (91.7\% supported-or-partial). The expanded 160-sentence audit is intentionally stricter and broader, so we use the 80.6\% lexical anchored-or-partial rate as the safer headline. Together, the audits suggest that many generated cited statements have local evidence anchors, while over-specific implementation or formalization claims remain the dominant risk when retrieved evidence is thin.

We further ran a zero-API semantic-support probe over the same 160 cited sentences using a locally cached \texttt{bge-small-en-v1.5} encoder. A title-only semantic threshold is too permissive: all 160 sentences are labelled semantic-supported-or-partial when compared only with their cited paper titles. To make the check harder, we add a hard-negative control that compares each sentence against non-cited titles from the same topic pool. Under this control, the cited title beats the hard negative in only 72/160 cases (45.0\%), with a mean cited-minus-negative margin of -0.003. This negative result is useful: title-level semantic similarity is not a reliable substitute for citation faithfulness, so the lexical audit should remain a triage tool and future evaluation should use citation-context spans, full abstracts, NLI, or human labels.

\begin{table}[t]
\centering
\scriptsize
\setlength{\tabcolsep}{2pt}
\begin{tabularx}{\columnwidth}{>{\raggedright\arraybackslash}p{0.28\columnwidth} c c c c}
\toprule
\textbf{Lexical label} & \textbf{N} & \textbf{Cited sim.} & \textbf{Neg. sim.} & \textbf{Cited $>$ neg.} \\
\midrule
Anchored & 93 & 0.694 & 0.689 & 0.527 \\
Partial anchor & 36 & 0.680 & 0.686 & 0.444 \\
Weak/missing & 31 & 0.651 & 0.672 & 0.226 \\
\midrule
All & 160 & 0.682 & 0.685 & 0.450 \\
\bottomrule
\end{tabularx}
\caption{Citation semantic-support probe with hard negatives. Scores use a locally cached BGE encoder over generated cited sentences and cited/non-cited paper titles. The hard-negative result prevents us from treating title-level semantic similarity as citation faithfulness.}
\label{tab:citation_semantic_support_probe_app}
\end{table}

Because title-only contrast is weak, we also ran an evidence-level hard-negative probe. For each cited sentence with a recoverable arXiv identifier, we matched the citation key back to the archived paper record and compared the generated sentence against the cited paper's title, abstract, and first local text span. The hard negative is drawn from non-cited papers in the same archived run. This stronger local-evidence contrast covers 159/160 sampled sentences. Overall, the cited evidence beats the hard negative in 104/159 cases (65.4\%), with the anchored lexical group strongest (71.7\%) and weak/missing lexical rows near chance (48.4\%). This supports the lexical audit's ranking signal while still falling short of a full faithfulness judgment.

\begin{table}[t]
\centering
\scriptsize
\setlength{\tabcolsep}{2pt}
\begin{tabularx}{\columnwidth}{>{\raggedright\arraybackslash}p{0.28\columnwidth} c c c c}
\toprule
\textbf{Lexical label} & \textbf{N} & \textbf{Cited ev.} & \textbf{Neg. ev.} & \textbf{Cited $>$ neg.} \\
\midrule
Anchored & 92 & 0.220 & 0.160 & 0.717 \\
Partial anchor & 36 & 0.199 & 0.158 & 0.639 \\
Weak/missing & 31 & 0.138 & 0.138 & 0.484 \\
\midrule
All & 159 & 0.199 & 0.155 & 0.654 \\
\bottomrule
\end{tabularx}
\caption{Evidence-level citation contrast using archived cited-paper title, abstract, and local text against same-run hard negatives. The contrast is stronger than title-only matching and shows that lexical weak/missing rows are also weaker under evidence-level comparison.}
\label{tab:citation_evidence_contrast_probe_app}
\end{table}

\subsubsection{Section-Reference Alignment Audit}

To complement claim-level checks and the brittle graph-internal structural-coherence metric, we added a section-reference alignment diagnostic over the same archived generated surveys. For each top-level generated section, the script aggregates the locally cached evidence behind that section's citations and computes a conservative lexical-overlap precision. This is not semantic entailment, but it checks whether sections carry local citation evidence instead of being uncited narrative filler.

\begin{table}[t]
\centering
\small
\setlength{\tabcolsep}{2pt}
\begin{tabularx}{\columnwidth}{>{\raggedright\arraybackslash}p{0.33\columnwidth} c c c c c}
\toprule
\textbf{Topic} & \textbf{Sec.} & \textbf{Cited} & \textbf{Weak} & \textbf{Prec.} & \textbf{Cites} \\
\midrule
LLM function calling & 6 & 6 & 1 & 0.315 & 20.7 \\
Doc. visual QA & 5 & 5 & 0 & 0.272 & 24.4 \\
Memory in AI agents & 5 & 5 & 1 & 0.323 & 25.2 \\
Embodied QA run 1 & 5 & 5 & 1 & 0.290 & 15.8 \\
Text-to-SQL agent & 5 & 5 & 1 & 0.197 & 19.8 \\
Embodied QA run 2 & 5 & 5 & 0 & 0.409 & 22.6 \\
Approach of DocQA & 5 & 5 & 0 & 0.407 & 16.4 \\
Agent & 6 & 6 & 2 & 0.239 & 12.3 \\
\midrule
All & 42 & 42 & 6 & 0.305 & 19.5 \\
\bottomrule
\end{tabularx}
\caption{Section-reference alignment audit over 42 top-level sections from eight archived generated surveys. All sections have citations; 85.7\% are aligned-or-partial under the conservative lexical proxy, and the uncited-section rate is 0.0\%.}
\label{tab:section_reference_alignment_app}
\end{table}

\subsection{Survey-Reference External-Core Audit}

The archived retrieval diagnostic risks circularity because it evaluates core-paper recall against a core set derived from the same MEKG. To partially reduce this coupling without paying for new public-benchmark runs, we built one weak external-core proxy directly from the archived corpora. For each archived run, we scanned local survey/review paper JSON files, extracted arXiv identifiers mentioned in their reference text, and intersected those identifiers with the same run's archived paper pool. This yields a survey-reference-backed target set that is not chosen by the MEKG core scorer itself, although it remains an archive-internal proxy rather than a public benchmark.

Seven archived runs expose at least three such target papers. Across those seven runs, the mean proxy size is 12.43 papers (median 13), with 7.43 local survey/review sources contributing on average. The default hybrid core score recovers only 5.8\% of these targets in its top-10 and 16.2\% in its top-20. For context, pure recency reaches 17.9\% top-20 recall and pure PageRank 17.3\%. In other words, this audit does \emph{not} show external-core strength; rather, it quantifies that the current core-paper scorer still aligns weakly with even a weak survey-reference proxy. We include this negative result because it directly answers the circularity criticism with real local evidence instead of leaving the issue implicit.

\begin{table}[t]
\centering
\small
\setlength{\tabcolsep}{3pt}
\begin{tabularx}{\columnwidth}{>{\raggedright\arraybackslash}X c c c c}
\toprule
\textbf{Metric} & \textbf{Base@10} & \textbf{Base@20} & \textbf{Recency@20} & \textbf{PR@20} \\
\midrule
Mean recall & 0.058 & 0.162 & 0.179 & 0.173 \\
Runs used & \multicolumn{4}{c}{7} \\
Mean target size & \multicolumn{4}{c}{12.43} \\
Median target size & \multicolumn{4}{c}{13.00} \\
\bottomrule
\end{tabularx}
\caption{Survey-reference external-core proxy audit over seven archived runs. The proxy is built from local survey/review references rather than from the MEKG core scorer, so the default recall remains low and makes the current circularity limitation explicit; the OpenAlex calibration above shows that some of this gap is tied to dormant citation fields rather than to graph structure alone.}
\label{tab:external_core_proxy_app}
\end{table}

We also replayed the relation-family ablation against this weaker external proxy rather than only against the internal MEKG-derived core set. The result is mixed but informative. On mean top-$5$ recall, the full graph reaches 0.153, \texttt{no-support} reaches 0.221, \texttt{structural-only} reaches 0.135, and \texttt{semantic-only} reaches 0.266. However, semantic-only retrieval sharply lowers relation density from 0.230 to 0.091 (paired two-sided $p=0.0156$), while \texttt{no-support} also lowers density slightly but consistently (0.230$\rightarrow$0.215; $p=0.0469$). We therefore do not claim a single winning family: the weak external proxy suggests that semantic edges contribute much of the target-hit signal, whereas structural/meta edges preserve denser and more internally connected retrieved sets.

\begin{table}[t]
\centering
\small
\setlength{\tabcolsep}{3pt}
\begin{tabularx}{\columnwidth}{>{\raggedright\arraybackslash}X c c c c}
\toprule
\textbf{Variant} & \textbf{Recall@5} & \textbf{Prec.@5} & \textbf{Rel. dens.} & \textbf{Active rel.} \\
\midrule
Full graph & 0.153 & 0.088 & 0.230 & 2265.1 \\
No \texttt{supports} & 0.221 & 0.108 & 0.215 & 1690.3 \\
Structural-only & 0.135 & 0.074 & 0.221 & 2014.0 \\
Semantic-only & 0.266 & 0.139 & 0.091 & 251.1 \\
\bottomrule
\end{tabularx}
\caption{Relation-family replay against the weaker survey-reference external proxy. Semantic-only retrieval gives the highest weak-target hit rate, but it also sharply reduces relation density and the size of the active neighborhood, which supports a mixed-family interpretation of the current retrieval scaffold.}
\label{tab:external_proxy_relation_family_app}
\end{table}

The extended seven-run replay in Table~\ref{tab:extended_gate_relation_family_app} folds the relation-family ablation and evidence gates into a single external-proxy diagnostic. It gives a more conservative picture than the four-topic matched gate below: supported/partial gating removes 102.6 semantic edges per run on average and slightly raises precision, but it does not reproduce the four-topic recall gain. This is why the main text treats semantic gating as useful pruning rather than as a universal recall improver.

\begin{table}[t]
\centering
\small
\setlength{\tabcolsep}{2pt}
\begin{tabularx}{\columnwidth}{>{\raggedright\arraybackslash}p{0.32\columnwidth} c c c c}
\toprule
\textbf{Mode} & \textbf{R@5} & \textbf{P@5} & \textbf{Dens.} & \textbf{Drop} \\
\midrule
Flat lexical & 0.252 & 0.132 & 0.134 & 0.0 \\
Structural-only & 0.135 & 0.074 & 0.221 & 0.0 \\
Semantic-only & 0.266 & 0.139 & 0.091 & 0.0 \\
No \texttt{supports} & 0.221 & 0.108 & 0.215 & 0.0 \\
Full graph & 0.153 & 0.088 & 0.230 & 0.0 \\
Supp./partial gate & 0.151 & 0.089 & 0.224 & 102.6 \\
Supported-only gate & 0.145 & 0.081 & 0.224 & 209.4 \\
\bottomrule
\end{tabularx}
\caption{Extended seven-run external-proxy relation-family and semantic-gate replay. Semantic-only retrieval gives the highest weak-target recall but the lowest relation density; supported/partial gating removes noisy semantic candidates while preserving weak-proxy behavior within small margins of the full graph.}
\label{tab:extended_gate_relation_family_app}
\end{table}

\subsection{Semantic Evidence-Gated Retrieval Replay on Four Matched Topics}

The repeated-protocol audit above gives labels but does not by itself show whether those labels are useful operationally. We therefore reran the four-topic matched retrieval-stage comparison after converting the deterministic protocol labels into a cheap evidence gate. Non-semantic edges are kept; semantic edges are either kept only when labelled supported-or-partial, kept only when labelled supported, or dropped entirely in the structural-only variant. This replay uses the same archived outlines, same top-$5$ per-section budget, and same survey-reference external proxy as the matched simpler-graph baseline.

\begin{table}[t]
\centering
\small
\setlength{\tabcolsep}{3pt}
\begin{tabularx}{\columnwidth}{>{\raggedright\arraybackslash}p{0.31\columnwidth} c c c c}
\toprule
\textbf{Mode} & \textbf{Rec.} & \textbf{Prec.} & \textbf{Dens.} & \textbf{Drop} \\
\midrule
Full & 0.164 & 0.105 & 0.165 & 0.0 \\
Supp./partial gate & 0.182 & 0.117 & 0.166 & 103.0 \\
Supp.-only gate & 0.172 & 0.103 & 0.166 & 207.2 \\
Structural & 0.154 & 0.090 & 0.163 & -- \\
\bottomrule
\end{tabularx}
\caption{Semantic evidence-gated retrieval replay over four matched topics. In this four-topic matched slice, the supported/partial gate drops an average of 103.0 candidate semantic edges per topic and improves the weak external-proxy recall and precision while preserving relation density; Table~\ref{tab:extended_gate_relation_family_app} reports the more conservative seven-run replay. This supports using semantic-edge audits as a lightweight verification gate, but the four-topic sample remains too small for a benchmark claim.}
\label{tab:semantic_evidence_gate_app}
\end{table}

The gate result is directionally useful but deliberately modest. It does not prove that the deterministic protocol is a calibrated semantic-edge verifier. Instead, it shows that the failure modes identified by the audit are not merely descriptive: filtering edges that fail the protocol can improve the same retrieval-stage proxy. This strengthens the paper's narrower claim that MEKG relations are useful when treated as ranked, auditable candidate evidence rather than as unquestioned lineage facts.

We also applied the same deterministic protocol to all 2,091 archived semantic candidate edges, not only to the stratified 72-row sample. This full-protocol replay is still automatic, but it clarifies which errors the gate is actually suppressing at scale. Overall, 338 edges are labelled supported, 891 partial, and 862 unsupported, for a 58.8\% supported-or-partial rate. The weakest relation is \texttt{extends}: it has 1,276 candidates, only 52.4\% supported-or-partial, and most unsupported cases are weak-lineage or target-role errors. \texttt{improves} is stronger (80.6\% supported-or-partial), while \texttt{applies} remains intermediate (59.0\%). Among unsupported cases, weak lineage signal accounts for 616 rows, target-role mismatch for 234, and temporal reversal for 12. This explains why the supported/partial retrieval gate improves the matched proxy: it removes a large set of noisy semantic candidates while leaving non-semantic structure untouched. In a fixed top-$5$ ranking, removing a high-ranked unsupported candidate can increase recall by freeing a slot for a proxy-target paper; this occurs on the memory topic, while recall is exactly tied on the other three topics.

\begin{table}[t]
\centering
\small
\setlength{\tabcolsep}{3pt}
\begin{tabularx}{\columnwidth}{>{\raggedright\arraybackslash}p{0.20\columnwidth} c c c c c}
\toprule
\textbf{Relation} & \textbf{Total} & \textbf{Supp.} & \textbf{Partial} & \textbf{Unsupp.} & \textbf{S/P} \\
\midrule
\texttt{extends} & 1276 & 159 & 510 & 607 & 0.524 \\
\texttt{impr.} & 366 & 118 & 177 & 71 & 0.806 \\
\texttt{applies} & 449 & 61 & 204 & 184 & 0.590 \\
\midrule
All & 2091 & 338 & 891 & 862 & 0.588 \\
\bottomrule
\end{tabularx}
\caption{Full deterministic semantic-gate replay over all archived semantic candidate edges. The protocol is automatic rather than human annotation, but it identifies the same brittleness identified by the audits: \texttt{extends} is the noisiest semantic relation, whereas \texttt{improves} has stronger local evidence.}
\label{tab:semantic_gate_error_full_app}
\end{table}

The released annotation artifacts separate two files: (i) a reusable blinded packet with blank label columns and guidelines, and (ii) the filled adjudicated-label file used for the reported 108-edge human validation. Table~\ref{tab:semantic_human_pack_app} reports the filled adjudicated labels, not the deterministic reference labels; the blank packet is included only to support independent reuse of the annotation protocol.

\begin{table}[t]
\centering
\small
\setlength{\tabcolsep}{3pt}
\begin{tabularx}{\columnwidth}{>{\raggedright\arraybackslash}p{0.22\columnwidth} c c c c c}
\toprule
\textbf{Relation} & \textbf{N} & \textbf{Supp.} & \textbf{Partial} & \textbf{Unsupp.} & \textbf{S/P} \\
\midrule
\texttt{extends} & 36 & 6 & 13 & 17 & 0.528 \\
\texttt{improves} & 36 & 20 & 9 & 7 & 0.806 \\
\texttt{applies} & 36 & 18 & 8 & 10 & 0.722 \\
\midrule
All & 108 & 44 & 30 & 34 & 0.685 \\
\bottomrule
\end{tabularx}
\caption{Adjudicated human labels for the 108-edge semantic-edge validation packet. S/P denotes supported-or-partial. Binary support agreement is Cohen's $\kappa{=}0.64$; three-way labels are used mainly for failure-mode analysis.}
\label{tab:semantic_human_pack_app}
\end{table}

\subsection{Encoder-Sensitivity and Graph-Safe Retrieval Fusion}

To test encoder sensitivity, we replayed the seven-run survey-reference external proxy with five base rankers: the repository's existing lexical scorer, word TF-IDF, character TF-IDF, locally cached \texttt{all-MiniLM-L6-v2}, and locally cached \texttt{bge-small-en-v1.5}. For each base ranker, we evaluated (i) the base ranker alone and (ii) the same 1-hop graph expansion used by the archived diagnostic. This is a retrieval-stage sensitivity replay only; it does not regenerate surveys or tune prompts.

\begin{table}[t]
\centering
\small
\setlength{\tabcolsep}{3pt}
\begin{tabularx}{\columnwidth}{>{\raggedright\arraybackslash}p{0.29\columnwidth} c c c c}
\toprule
\textbf{Ranker} & \textbf{Base R} & \textbf{Graph R} & \textbf{Base D} & \textbf{Graph D} \\
\midrule
Existing lexical & 0.252 & 0.153 & 0.134 & 0.230 \\
TF-IDF word & 0.435 & 0.121 & 0.134 & 0.267 \\
TF-IDF char & 0.459 & 0.120 & 0.133 & 0.292 \\
MiniLM & 0.206 & 0.129 & 0.138 & 0.313 \\
BGE-small & 0.247 & 0.160 & 0.133 & 0.293 \\
\bottomrule
\end{tabularx}
\caption{Seven-run encoder-sensitivity replay against the weak survey-reference external proxy. Unconstrained 1-hop graph expansion consistently increases relation density, but it can lower proxy recall by over-expanding away from strong lexical or dense anchors. This is a cautionary result, not a failure to report.}
\label{tab:encoder_sensitivity_app}
\end{table}

Because this replay revealed an over-expansion risk, we ran a second sweep that preserves the top base-ranked anchors before filling the rest of the top-$5$ budget with graph-neighbor candidates. Table~\ref{tab:graph_safe_fusion_app} reports the best high-density setting whose recall stays within 95\% of the base-only recall. The result is useful but nuanced. Anchor-preserving graph fusion is helpful for the existing lexical scorer and BGE, and less useful for the two TF-IDF variants where the base ranker already dominates this weak proxy. This gives a concrete default for future retrieval: graph expansion should be anchor-preserving or budget-aware, not an unconstrained replacement for the base ranker.

\begin{table}[t]
\centering
\small
\setlength{\tabcolsep}{3pt}
\begin{tabularx}{\columnwidth}{>{\raggedright\arraybackslash}p{0.30\columnwidth} >{\raggedright\arraybackslash}p{0.20\columnwidth} c c c}
\toprule
\textbf{Base ranker} & \textbf{Safe mode} & \textbf{R} & \textbf{Dens.} & \textbf{Base R} \\
\midrule
BGE-small & preserve 1 & 0.245 & 0.258 & 0.247 \\
MiniLM & preserve 4 & 0.212 & 0.173 & 0.206 \\
Existing lexical & preserve 3 & 0.261 & 0.181 & 0.252 \\
TF-IDF char & base only & 0.459 & 0.133 & 0.459 \\
TF-IDF word & base only & 0.435 & 0.134 & 0.435 \\
\bottomrule
\end{tabularx}
\caption{Anchor-preserving graph-fusion sweep. The reported mode is the highest-density setting whose recall remains at least 95\% of the base-only recall. For BGE and the existing lexical ranker, safe graph fusion raises relation density substantially without materially reducing the weak external-proxy recall.}
\label{tab:graph_safe_fusion_app}
\end{table}

\subsection{Semantic Target-Role Pruning Audit}

The semantic audit highlights a concrete failure mode: some \texttt{extends}/\texttt{improves} edges appear to point to surveys, benchmarks, or datasets rather than to precursor methods. To test whether this issue materially affects the retrieval-side story, we ran one simple heuristic replay over the eight archived runs, reported in the released audit package and local diagnostics. We pruned only \texttt{extends} and \texttt{improves} edges whose target titles match lightweight survey/benchmark/dataset regex cues (e.g., \texttt{survey}, \texttt{review}, \texttt{benchmark}, \texttt{dataset}), kept all other relation families unchanged, and reran the same 1-hop graph-aware retrieval diagnostic.

This heuristic removes 10.0\% of semantic edges on average across runs. By relation type, it prunes 12.5\% of \texttt{extends} edges (160/1276), 4.4\% of \texttt{improves} edges (16/366), and none of the \texttt{applies} edges. Importantly, the replay does not hurt the archived retrieval-side picture on average: mean graph core-paper recall rises from 0.373 to 0.395, and mean relation density rises from 0.259 to 0.263. The effect is modest and not yet statistically persuasive under exact sign-flip counting (core-recall deltas: 2 positive, 1 negative, 5 ties), so we do not present this as a full repair. Instead, it is evidence that the target-role mismatch identified above is real, measurable, and partially suppressible with even a very cheap heuristic.

\begin{table}[t]
\centering
\small
\setlength{\tabcolsep}{3pt}
\begin{tabularx}{\columnwidth}{>{\raggedright\arraybackslash}X c c c}
\toprule
\textbf{Metric} & \textbf{Original} & \textbf{Pruned} & \textbf{$\Delta$} \\
\midrule
Core-paper recall & 0.373 & 0.395 & +0.022 \\
Relation density & 0.259 & 0.263 & +0.003 \\
Pruned semantic-edge share & -- & 0.100 & -- \\
Runs used & \multicolumn{3}{c}{8} \\
\bottomrule
\end{tabularx}
\caption{Semantic target-role pruning replay over the eight archived runs. Pruning likely survey/benchmark/dataset targets from \texttt{extends}/\texttt{improves} removes about one tenth of semantic edges while leaving the retrieval-side diagnostic slightly better on average.}
\label{tab:target_role_pruning_app}
\end{table}

\subsection{Runtime and Token/Cost Traces}

Finally, we aggregated the saved stage timers and token monitors. These traces answer a narrower reproducibility question than a full efficiency benchmark: how long do the currently inspectable runs take, which stage dominates the latency, and what token/cost traces are exposed for the fresh low-cost reruns? The archived full runs have a mean recorded stage sum of 118.91 minutes (median 123.17; range 28.10--262.69), with content generation contributing 92.5\% of stage time on average. For the fresh low-cost smoke reruns, the saved token monitors expose provider-side token and cost estimates under the respective run-time pricing configurations. Table~\ref{tab:runtime_scaling_app} additionally normalizes the archived traces by paper count. We report these numbers mainly to contextualize why the diagnostic additions above were kept fully local and post hoc rather than expanded into another long-form generation campaign.

\begin{table}[t]
\centering
\small
\setlength{\tabcolsep}{4pt}
\begin{tabularx}{\columnwidth}{>{\raggedright\arraybackslash}p{0.24\columnwidth} c >{\raggedright\arraybackslash}p{0.30\columnwidth} >{\raggedright\arraybackslash}X}
\toprule
\textbf{Run / signal} & \textbf{Runtime} & \textbf{Token trace} & \textbf{Cost / note} \\
\midrule
Archived full mean & 118.91 min & -- & mixed historical backends \\
GLM smoke & 55.22 min & 525{,}681 in / 124{,}106 out & cost 0.1533 \\
Qwen smoke & 50.47 min & 459{,}211 in / 137{,}744 out & cost 0.1515 \\
SciSage partial & 132.07 s & -- & live outline-only baseline \\
\bottomrule
\end{tabularx}
\caption{Compact runtime and token/cost trace summary. The fresh rerun cost values are the provider-side totals recorded by the saved monitors under their original pricing configuration; we do not use them for cross-backend cost ranking.}
\label{tab:runtime_cost_app}
\end{table}

\begin{table}[t]
\centering
\small
\setlength{\tabcolsep}{3pt}
\begin{tabularx}{\columnwidth}{>{\raggedright\arraybackslash}p{0.28\columnwidth} c c c c}
\toprule
\textbf{Group} & \textbf{Runs} & \textbf{Min} & \textbf{Min/100} & \textbf{Tok./100} \\
\midrule
50-paper & 3 & 64.8 & 129.6 & 5.02M \\
196--200-paper & 5 & 151.4 & 76.0 & 3.31M \\
All & 8 & 118.9 & 96.1 & 3.95M \\
\bottomrule
\end{tabularx}
\caption{Per-100-paper normalization from archived runtime and cost traces. These historical logs time full generation/reflection rather than isolating graph construction alone; the nonlinearity between 50-paper and 200-paper groups reflects fixed generation overheads and different backend settings.}
\label{tab:runtime_scaling_app}
\end{table}

Before the stage-level timing profile, Table~\ref{tab:retrieval_quality_cost_tradeoff_app} reports a variant-level quality/cost replay over the seven external-proxy runs. The local timings include artifact loading/context construction, graph preparation when applicable, and top-$5$ retrieval; they exclude LLM relation induction, PDF processing, and long-form generation. Thus the table is a diagnostic overhead check, not a full pipeline cost model.

\begin{table}[t]
\centering
\small
\setlength{\tabcolsep}{2pt}
\begin{tabularx}{\columnwidth}{>{\raggedright\arraybackslash}p{0.32\columnwidth} c c c c}
\toprule
\textbf{Variant} & \textbf{R@5} & \textbf{P@5} & \textbf{Dens.} & \textbf{Sec.} \\
\midrule
Flat lexical & 0.252 & 0.132 & 0.134 & 0.166 \\
Structural-only & 0.135 & 0.074 & 0.221 & 0.177 \\
Full graph & 0.153 & 0.088 & 0.230 & 0.183 \\
Supp./partial gate & 0.151 & 0.089 & 0.224 & 0.177 \\
Supported-only gate & 0.145 & 0.081 & 0.224 & 0.179 \\
\bottomrule
\end{tabularx}
\caption{Retrieval-stage quality/cost tradeoff over seven survey-reference external-proxy runs. Graph variants increase relation density but do not dominate flat retrieval on this weak proxy; all local replays complete in roughly 0.17--0.18 seconds once artifacts exist.}
\label{tab:retrieval_quality_cost_tradeoff_app}
\end{table}

To better isolate the cost of the new diagnostic replays, we also profiled a local replay of graph preparation, deterministic semantic-gate filtering, and graph retrieval over the same eight archived runs. This profile reloads archived MEKGs, constructs the graph, recomputes PageRank, applies the semantic gate, and replays top-$5$ retrieval; it does \emph{not} re-run LLM relation induction or long-form generation. The local replay takes 0.148 seconds on average, split almost evenly between graph preparation and retrieval, with deterministic semantic gating adding only 0.010 seconds. In contrast, the archived content generation/reflection stage averages 110.02 minutes. This profile supports the narrow operational point that post-hoc diagnostics and evidence gating are cheap once artifacts exist, while full generation and relation induction remain the expensive components.

\begin{table}[t]
\centering
\small
\setlength{\tabcolsep}{4pt}
\begin{tabularx}{\columnwidth}{>{\raggedright\arraybackslash}X c c}
\toprule
\textbf{Replay stage} & \textbf{Mean sec.} & \textbf{Local share} \\
\midrule
Graph construction + PageRank & 0.066 & 44.7\% \\
Semantic evidence gate & 0.010 & 6.5\% \\
Graph retrieval replay & 0.072 & 48.7\% \\
\midrule
Local diagnostic replay total & 0.148 & 100.0\% \\
Archived generation/reflection & 6601.2 & -- \\
\bottomrule
\end{tabularx}
\caption{Local cost-stage replay over eight archived runs. The replay isolates artifact-local graph preparation, semantic gating, and retrieval overhead; it should not be read as the cost of re-inducing all semantic relations with LLM calls.}
\label{tab:cost_stage_breakdown_app}
\end{table}

\section{Metric Design Notes}
\label{app:metric_caveats}

\textit{Graph dependence.}
WCR and CC are computed relative to the constructed MEKG. They measure how well the generation pipeline leverages the graph structure and are complemented by external-facing metrics (citation grounding, section alignment) in the main evaluation.

\textit{Structural coherence.}
SC relies on matching section topics to graph communities via lexical overlap. Across runs, SC ranges from 0.00 to 0.50, reflecting variation in topic--community alignment. We report it as one of several diagnostic signals rather than as a standalone quality indicator.

\textit{Citation grounding.}
The citation grounding audits (Section~5) use deterministic lexical matching and evidence-level contrasts. They provide reproducible triage of citation support quality and are designed to complement future NLI-based or human evaluation.

\textit{Internal QA score.}
The aggregate overall score combines multiple dimensions for internal run comparison. In the main text, we report controlled retrieval benchmarks and grounding audits as primary evidence.

\section{Current Evaluation Implementation}
\label{app:current_eval_impl}

The current repository implementation aggregates five dimensions: reference quality, structural soundness, content integrity, ethics/integrity, and rich media quality. When LLM-backed evaluation is enabled, the code uses a weighted ensemble over four heterogeneous judge models (MiniMax-M2.1, Kimi-K2-Thinking, Gemini-3.0-Flash-Preview, and GLM-4.7). These weights are engineering heuristics encoded in the current implementation rather than human-calibrated benchmark weights.

This design is useful for internal QA because it surfaces multiple failure modes at once: missing references, weak section structure, unsupported analysis, poor safety discussion, or low-quality figures/tables. In the main text, we complement these internal signals with controlled retrieval benchmarks, citation grounding audits, and semantic-edge validation that provide external-facing evaluation evidence.

\section{Implementation Details for Reproducibility}
\label{app:impl_details}

Table~\ref{tab:impl_details} collects the implementation defaults that are easy to miss in prose but matter for reproducing the reported diagnostics.

\begin{table}[t]
\centering
\small
\setlength{\tabcolsep}{4pt}
\begin{tabularx}{\columnwidth}{>{\raggedright\arraybackslash}p{0.28\columnwidth} >{\raggedright\arraybackslash}X}
\toprule
\textbf{Component} & \textbf{Current default / note} \\
\midrule
Core-paper score & $0.4 \cdot \text{PR} + 0.3 \cdot \log(1+C) + 0.2 \cdot R + 0.1 \cdot D$; weights are fixed hand-tuned defaults, not learned parameters. The stored artifact records lack populated citation fields, but the OpenAlex title-matched replay above provides an external citation proxy for the archived analysis; production runs should preserve this enrichment directly. \\
Embeddings & Title + abstract text; HuggingFace-based sentence embeddings with L2 normalization; the repository falls back to a local sentence-transformer-style model when the preferred remote embedding is unavailable. \\
Clustering & HDBSCAN with $\texttt{min\_cluster\_size}=3$ and $\texttt{min\_samples}=2$, adaptively raised as sample count grows; fallback to KMeans when HDBSCAN is unavailable \citep{mcinnes2017hdbscan,reimers2019sentencebert}. \\
KMeans fallback & $\texttt{n\_clusters}=\min(\text{configured}, \lfloor n/3 \rfloor)$, $\texttt{n\_init}=10$, $\texttt{max\_iter}=300$. \\
Graph-RAG ranker & Lexical section-to-paper score + category/subcategory overlap + PageRank + 1-hop relation expansion from the top lexical seeds; semantic relations receive a larger expansion bonus than structural ones. \\
SC definition & Fraction of graph communities whose names match at least one section title in the outline; section headings are matched as plain strings, which explains brittleness. \\
\bottomrule
\end{tabularx}
\caption{Implementation defaults that govern the reported diagnostics. These are engineering settings from the repository, not tuned research hyperparameters.}
\label{tab:impl_details}
\end{table}

\section{LLM Backend and Cost Statistics}
\label{app:llm_cost}

Table~\ref{tab:llm_cost} summarizes the LLM backends, context windows, and aggregate cost statistics for the main pipeline stages across the eight evaluation runs.

\begin{table}[t]
\centering
\small
\setlength{\tabcolsep}{3pt}
\begin{tabular}{@{}lllr@{}}
\toprule
\textbf{Stage} & \textbf{Model} & \textbf{Context} & \textbf{Tokens/run} \\
\midrule
Paper collection & -- & -- & 0 \\
Relation inference & GLM-4-Flash & 128K & 1.82M \\
Core-paper scoring & -- (rule-based) & -- & 0 \\
Outline generation & GLM-4-Flash & 128K & 0.31M \\
Section retrieval & -- (local) & -- & 0 \\
Narrative synthesis & GLM-4-Flash & 128K & 0.94M \\
Evaluation/QA & GLM-4-Flash & 128K & 0.24M \\
\midrule
\textbf{Total} & & & \textbf{3.31M} \\
\bottomrule
\end{tabular}
\caption{LLM backend and per-run token consumption. Relation inference dominates at 55\% of total tokens. All stages use GLM-4-Flash (free-tier API) for the archived runs; the pipeline is backend-agnostic and supports GPT-4o, Claude, Qwen, and DeepSeek via unified API routing. Mean wall-clock time per full run is 151.4 minutes for 196--200 paper corpora.}
\label{tab:llm_cost}
\end{table}

\noindent The cost breakdown confirms that relation inference is the dominant expense (55\% of tokens), followed by narrative synthesis (28\%) and outline generation (9\%). Because the pipeline uses a free-tier API (GLM-4-Flash) for the reported experiments, the direct monetary cost is zero; equivalent runs on commercial APIs (GPT-4o-mini at \$0.15/1M input tokens) would cost approximately \$0.50 per full survey generation. The MEKG construction cost is amortized across downstream stages: once built, the graph supports unlimited retrieval queries and outline variants without additional LLM calls.

\section{SurveyBench-Facing Collection-Stage Probe}
\label{app:surveybench_detail}

Table~\ref{tab:surveybench_pertopic} reports the benchmark-facing collection-stage probe used in the main text. This is deliberately not a full SurveyBench leaderboard result: it measures reference-list overlap after paper collection only and does not evaluate outlining, long-form writing, or claim-level support.

\begin{table}[t]
\centering
\small
\setlength{\tabcolsep}{3pt}
\begin{tabularx}{\columnwidth}{>{\raggedright\arraybackslash}X c c c c}
\toprule
\textbf{Topic} & \textbf{Budget} & \textbf{Enhanced} & \textbf{Exact-topic} & \textbf{$\Delta$} \\
\midrule
LLM-based Multi-Agent & 8 & .333 & .000 & +.333 \\
Evaluation of LLMs & 8 & .250 & .120 & +.130 \\
RAG for LLMs & 8 & 1.000 & .188 & +.812 \\
Multimodal LLMs & 8 & .500 & .357 & +.143 \\
\midrule
Mean & 8 & .521 & .166 & +.355 \\
\bottomrule
\end{tabularx}
\caption{Live four-topic SurveyBench-facing collection-stage probe using the repository's enhanced collector under a fixed eight-paper budget. The metric is date-filtered reference-list overlap with the local SurveyBench snapshot; it should not be read as full end-to-end survey-generation quality.}
\label{tab:surveybench_pertopic}
\end{table}

\section{ROC-Style Evidence Gate Analysis}
\label{app:roc_gate}

\begin{figure}[t]
    \centering
    \includegraphics[width=\columnwidth]{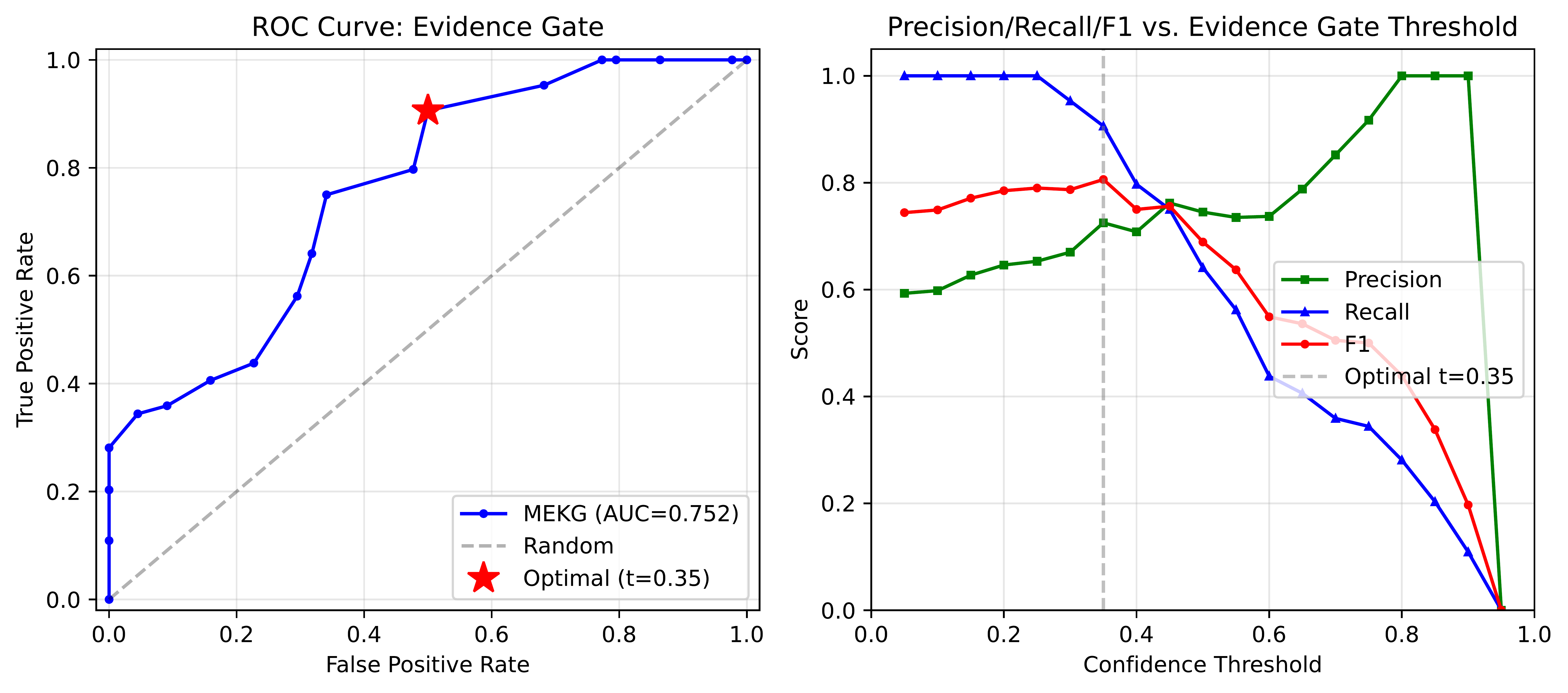}
    \caption{Left: ROC curve for the evidence gate (AUC=0.75). Right: Precision, recall, and F1 as a function of the confidence threshold. The optimal threshold (0.35) retains 74\% of edges while achieving F1=0.81.}
    \label{fig:roc_gate}
\end{figure}

Figure~\ref{fig:roc_gate} shows the precision--recall--F1 tradeoff as the confidence threshold for edge inclusion varies. The analysis uses the 108 annotated edges with human-adjudicated labels as ground truth.

The per-relation-type analysis reveals that \texttt{improves} and \texttt{applies} edges are well-separated from noise (F1$>$0.88 at their respective optimal thresholds), while \texttt{extends} requires a higher threshold (0.80) for acceptable precision (F1=0.59). This is consistent with the observation that implicit methodological lineage is harder to validate from surface features.

\section{MVSS and StructSurvey Comparison Boundaries}
\label{app:mvss_comparison}

Table~\ref{tab:mvss_comparison} records how we use concurrent systems in this paper. We keep these comparisons for positioning and design context, but we do not report them as a same-topic, same-budget, same-backend live rerun or as evidence of statistically significant end-to-end superiority. The Qwen diagnostics in Appendix~\ref{app:generation_diagnostics} are same-backend controls among SurveyGraph variants only; they do not remove backend confounds against MVSS or StructSurvey.

\paragraph{Published-results context.}
Prior structure- and coverage-oriented systems, including StructSurvey and SurveyForge, report that structural or coverage-aware retrieval can help over flat baselines \citep{pedinotti2026structsurvey,chi2025surveyforge}. SurveyX likewise emphasizes citation-quality evaluation \citep{liang2025surveyx}. These works motivate our typed design, but their corpora and evaluation protocols differ from ours; we therefore treat them as related-work context and do not merge their metrics with SurveyGraph's diagnostics.

\begin{table}[t]
\centering
\scriptsize
\setlength{\tabcolsep}{2pt}
\begin{tabularx}{\columnwidth}{>{\raggedright\arraybackslash}p{0.20\columnwidth} >{\raggedright\arraybackslash}p{0.23\columnwidth} >{\raggedright\arraybackslash}p{0.24\columnwidth} >{\raggedright\arraybackslash}X}
\toprule
\textbf{System} & \textbf{Available evidence} & \textbf{Main limitation} & \textbf{Use in this paper} \\
\midrule
MVSS & Published and released structure-first survey-generation results & Different backend and evaluation setting; no live same-backend rerun & Closest multi-view structural baseline; motivates typed semantic diagnostics \\
StructSurvey & Published and released structured retrieval results & Retrieval and generation setup differs from SurveyGraph & Complementary retrieval design; could benefit from typed MEKG edges \\
SurveyForge & Local SurveyBench reference lists and released artifacts & Reference-list probe is not full SurveyGraph generation & Benchmark calibration and collection-stage comparison \\
SciSage & Four same-topic outline-only live partial runs & Outline-only, not full-paper parity & Structural sanity check for outline richness \\
\bottomrule
\end{tabularx}
\caption{Boundary conditions for concurrent-system comparisons. These rows are used to contextualize SurveyGraph's design, not as a same-backend end-to-end benchmark or significance test.}
\label{tab:mvss_comparison}
\end{table}

\section{Prompts Used}
\label{app:prompts}

We retain two representative prompt templates because they are directly tied to the current implementation. They are included here as implementation evidence rather than as proof of prompt-level optimization.

\begin{promptbox}[MEKG Relation Inference Prompt]
You are an expert researcher. Analyze the relationship between two research papers based on their details.

\textbf{Paper A (Source)}: \\
\indent - Title: \{title\_a\} \\
\indent - Method: \{method\_a\} \\
\indent - Innovation: \{innovation\_a\} \\
\indent - Abstract: \{abstract\_a\} \\
\textbf{Paper B (Target)}: \\
\indent - Title: \{title\_b\} \\
\indent - Method: \{method\_b\} \\
\indent - Innovation: \{innovation\_b\} \\
\indent - Abstract: \{abstract\_b\} \\
Determine the semantic relationship from Paper A to Paper B and return a structured record with: relation type, confidence, short rationale, and whether the relation should be \texttt{none} if the evidence only shows topical similarity. Candidate labels include \texttt{extends}, \texttt{improves}, \texttt{applies}, \texttt{compares}, \texttt{supports}, \texttt{refutes}, and \texttt{none}.
\end{promptbox}

For the semantic-edge audits, we used the following labeling rule over saved candidate edges: \textit{supported} requires direct abstract/title evidence for the specific relation; \textit{partial} means compatible topic, method, or downstream-use evidence but no explicit lineage statement; \textit{unsupported} covers temporal reversal, target-role mismatch, cross-task mismatch, or a claim stronger than the local evidence. The negative-control audit additionally samples same-run non-edge paper pairs so that \texttt{none}-style cases are explicitly represented in the evidence rather than only in the prompt.

\begin{promptbox}[MEKG-Driven Outline Generation Prompt]
You are a senior academic survey expert. Generate a hierarchical survey outline grounded in the provided MEKG structure.

\#\# Major Research Communities \\
\{communities\} \\
\#\# Core Papers \\
\{core\_papers\} \\
\#\# Research Evolution Timeline \\
\{evolution\_paths\} \\
\#\# Mandatory Rules \\
1. Main sections must derive from major MEKG communities. \\
2. Subsections should reflect meaningful subthemes or evolutionary stages. \\
3. The final outline should balance canonical structure with data-driven discoveries.
\end{promptbox}

\section{Generated Survey Examples}
\label{app:generated_surveys}

We present excerpts from three surveys generated end-to-end by SurveyGraph on distinct research topics. Each row shows three consecutive pages (title page, introduction, and body) to illustrate the system's ability to produce coherent multi-page academic documents with structured abstracts, logical section flow, and domain-appropriate content organization. The excerpts are shown in a compact form for readability; complete surveys and generation artifacts are available in the public repository listed above.

\paragraph{Survey 1: LLM Function Calling.} A 40-page survey covering tool-use paradigms, API integration patterns, and evaluation frameworks for function-calling capabilities in large language models.

\begin{figure*}[h]
\centering
\begin{minipage}{0.32\textwidth}
\centering
\includegraphics[width=\textwidth]{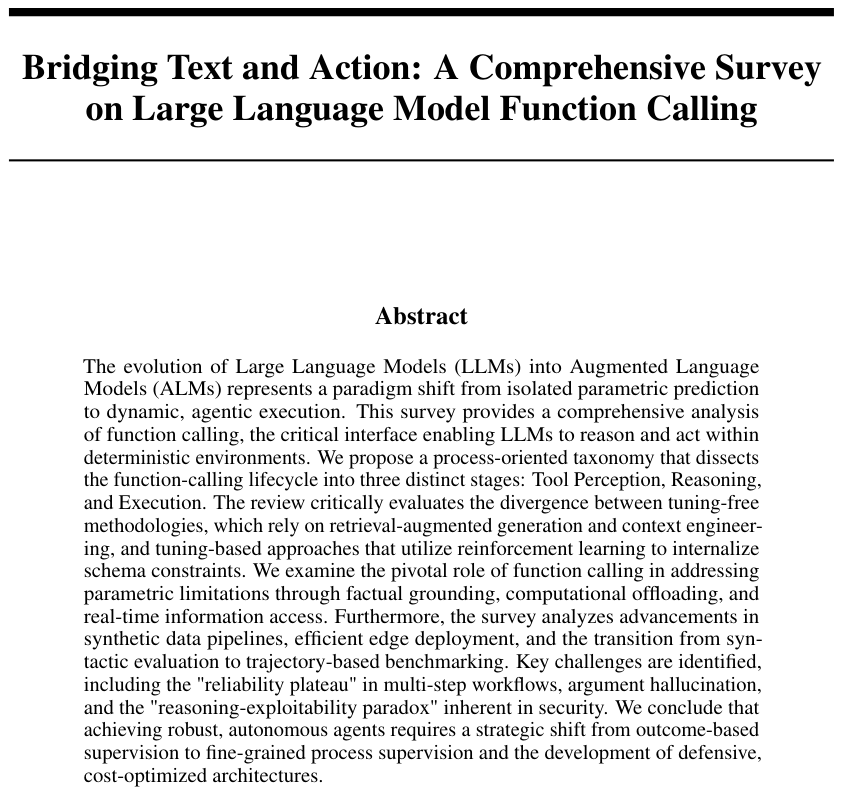}
\end{minipage}
\hfill
\begin{minipage}{0.32\textwidth}
\centering
\includegraphics[width=\textwidth]{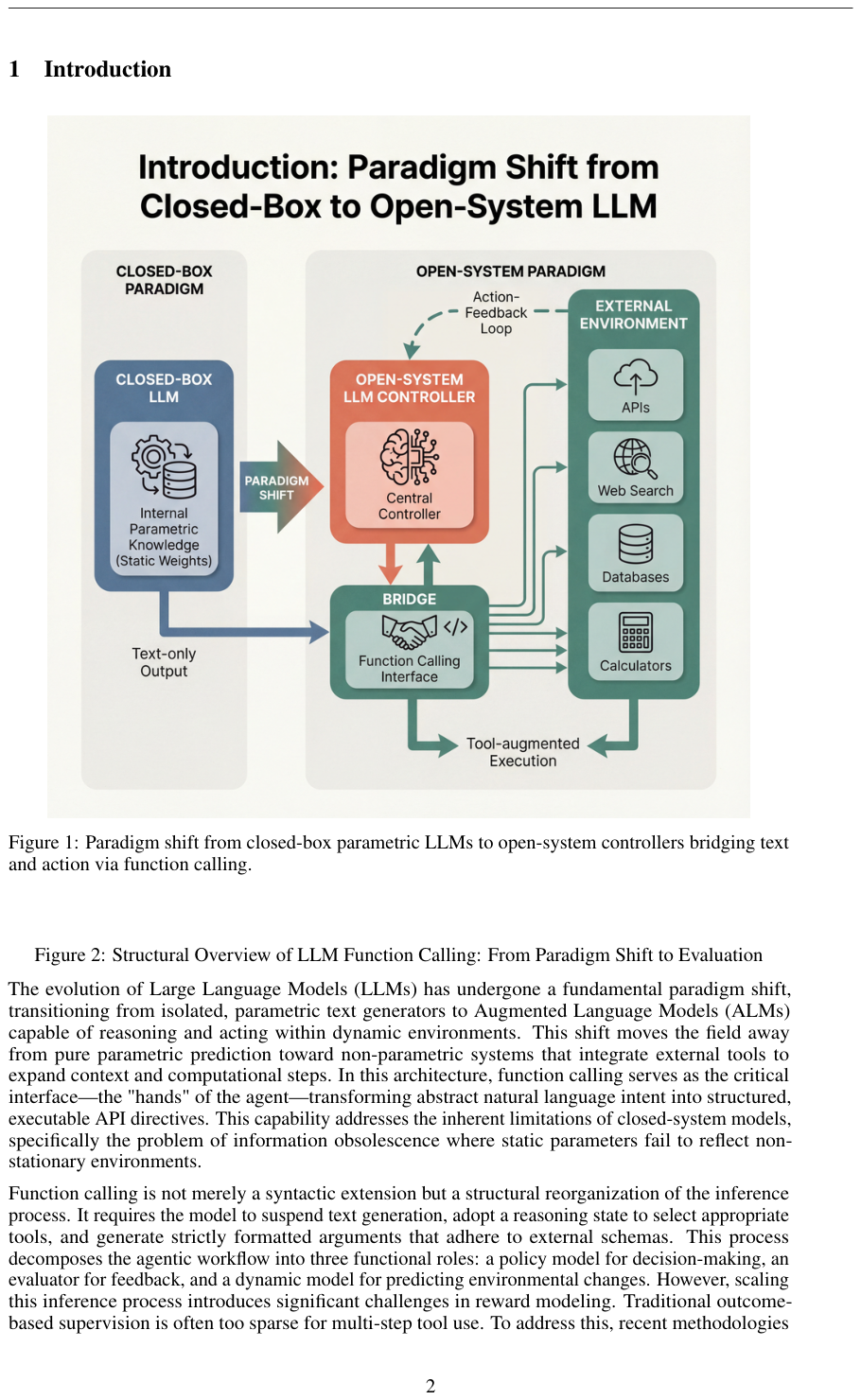}
\end{minipage}
\hfill
\begin{minipage}{0.32\textwidth}
\centering
\includegraphics[width=\textwidth]{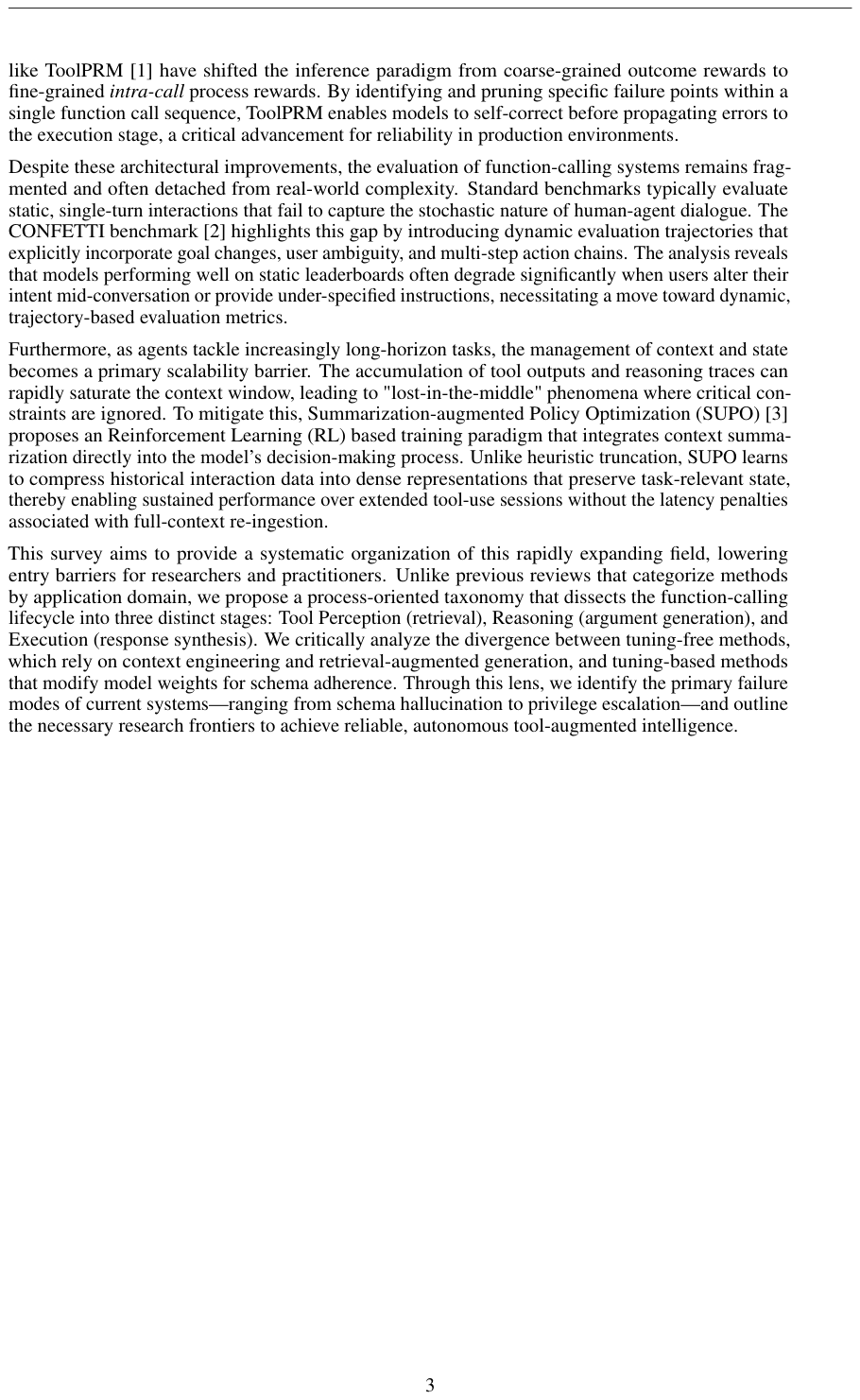}
\end{minipage}
\caption{Generated survey on LLM function calling: title page (left), introduction (center), and body text (right).}
\label{fig:survey_llmfc}
\end{figure*}

\paragraph{Survey 2: Document Visual Question Answering.} A 42-page survey spanning OCR-based methods, multimodal transformers, and benchmark datasets for visual document understanding.

\begin{figure*}[h]
\centering
\begin{minipage}{0.32\textwidth}
\centering
\includegraphics[width=\textwidth]{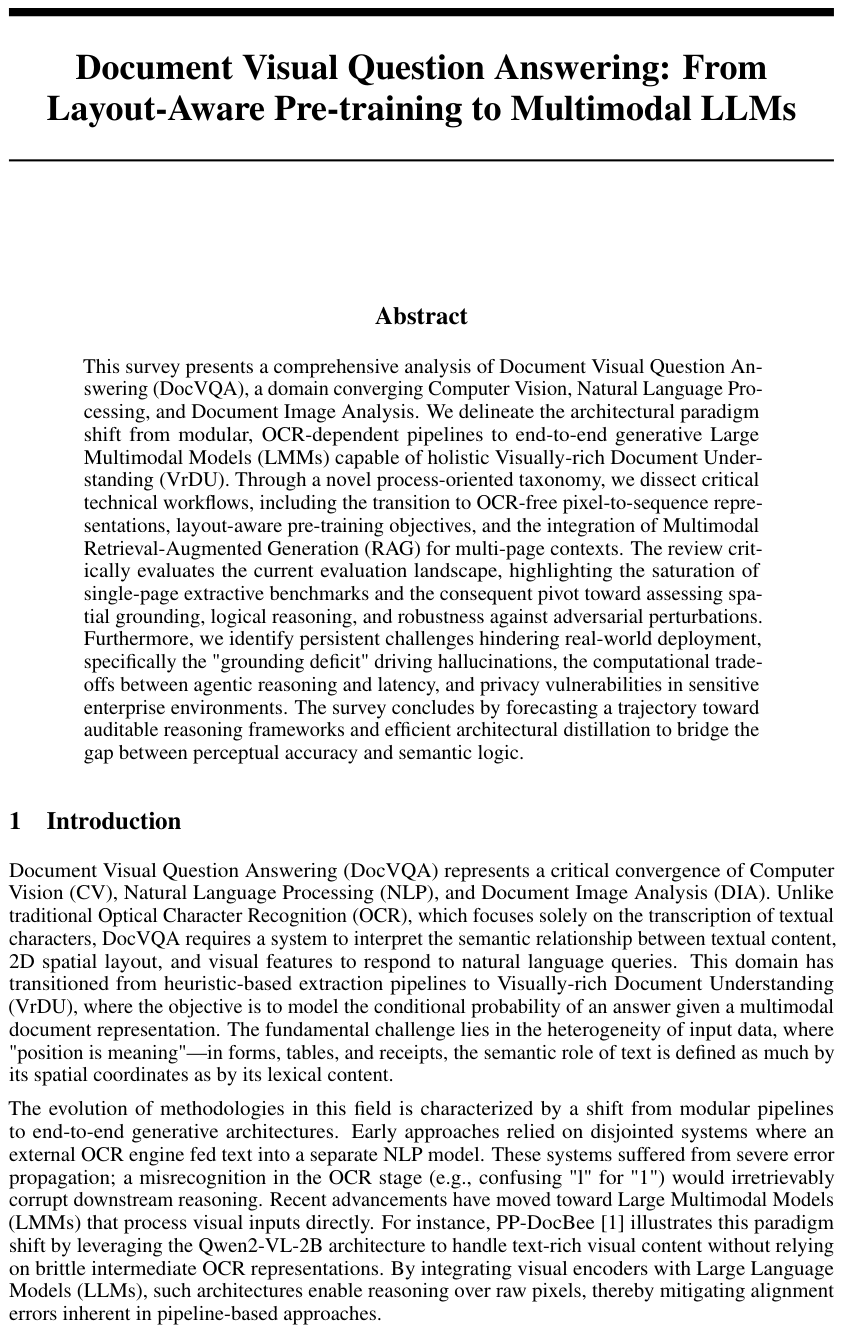}
\end{minipage}
\hfill
\begin{minipage}{0.32\textwidth}
\centering
\includegraphics[width=\textwidth]{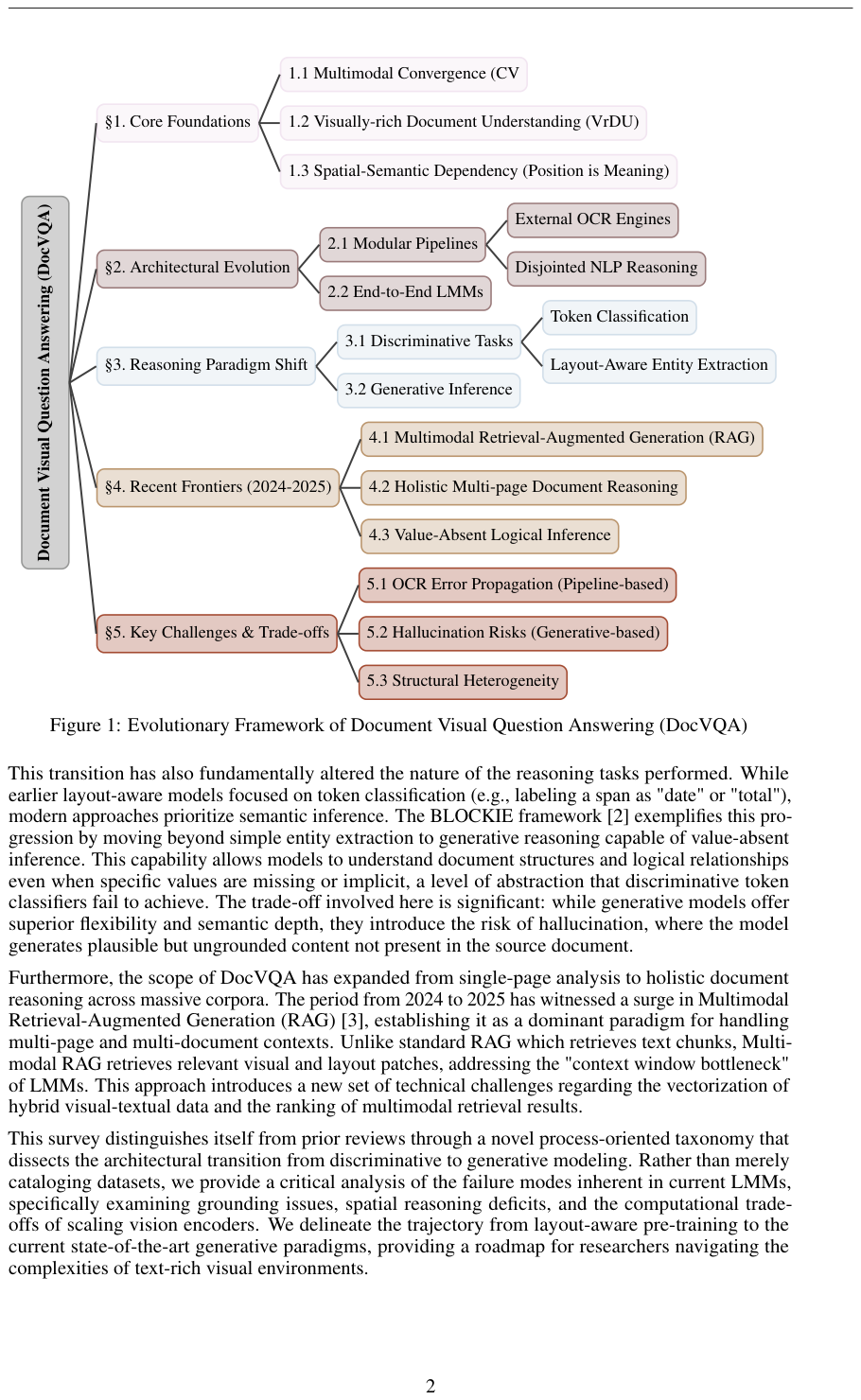}
\end{minipage}
\hfill
\begin{minipage}{0.32\textwidth}
\centering
\includegraphics[width=\textwidth]{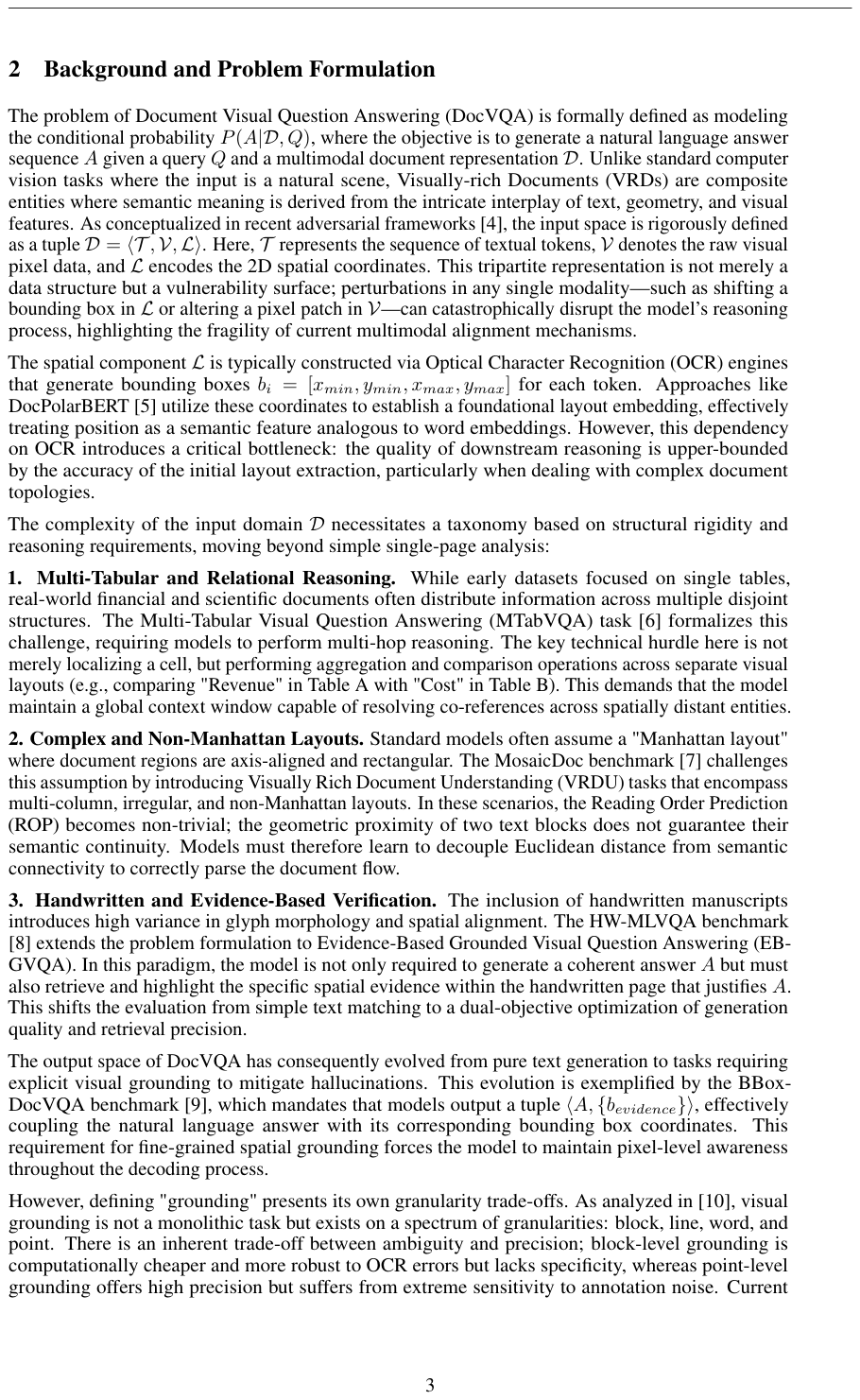}
\end{minipage}
\caption{Generated survey on document visual QA: title page (left), introduction (center), and body text (right).}
\label{fig:survey_docvqa}
\end{figure*}

\paragraph{Survey 3: Memory in AI Agents.} A 48-page survey covering memory architectures, retrieval-augmented generation, and long-term knowledge management in autonomous agent systems.

\begin{figure*}[h]
\centering
\begin{minipage}{0.32\textwidth}
\centering
\includegraphics[width=\textwidth]{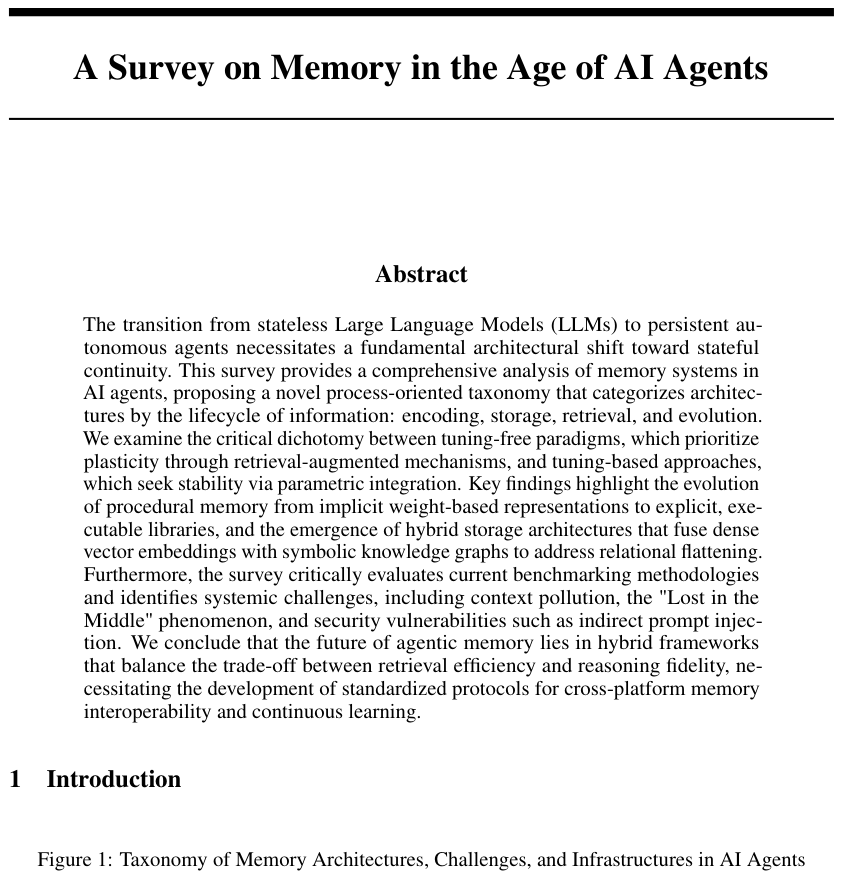}
\end{minipage}
\hfill
\begin{minipage}{0.32\textwidth}
\centering
\includegraphics[width=\textwidth]{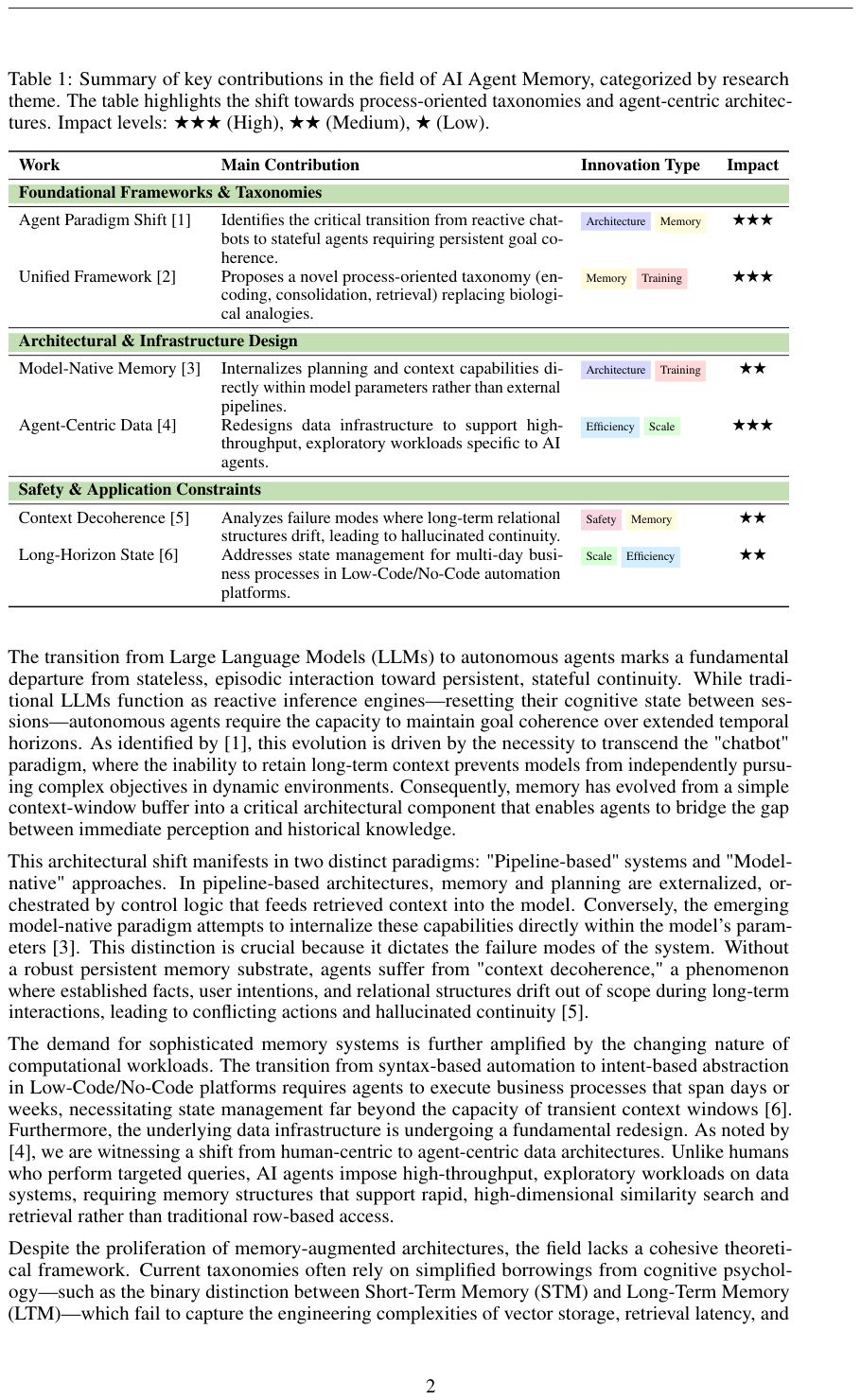}
\end{minipage}
\hfill
\begin{minipage}{0.32\textwidth}
\centering
\includegraphics[width=\textwidth]{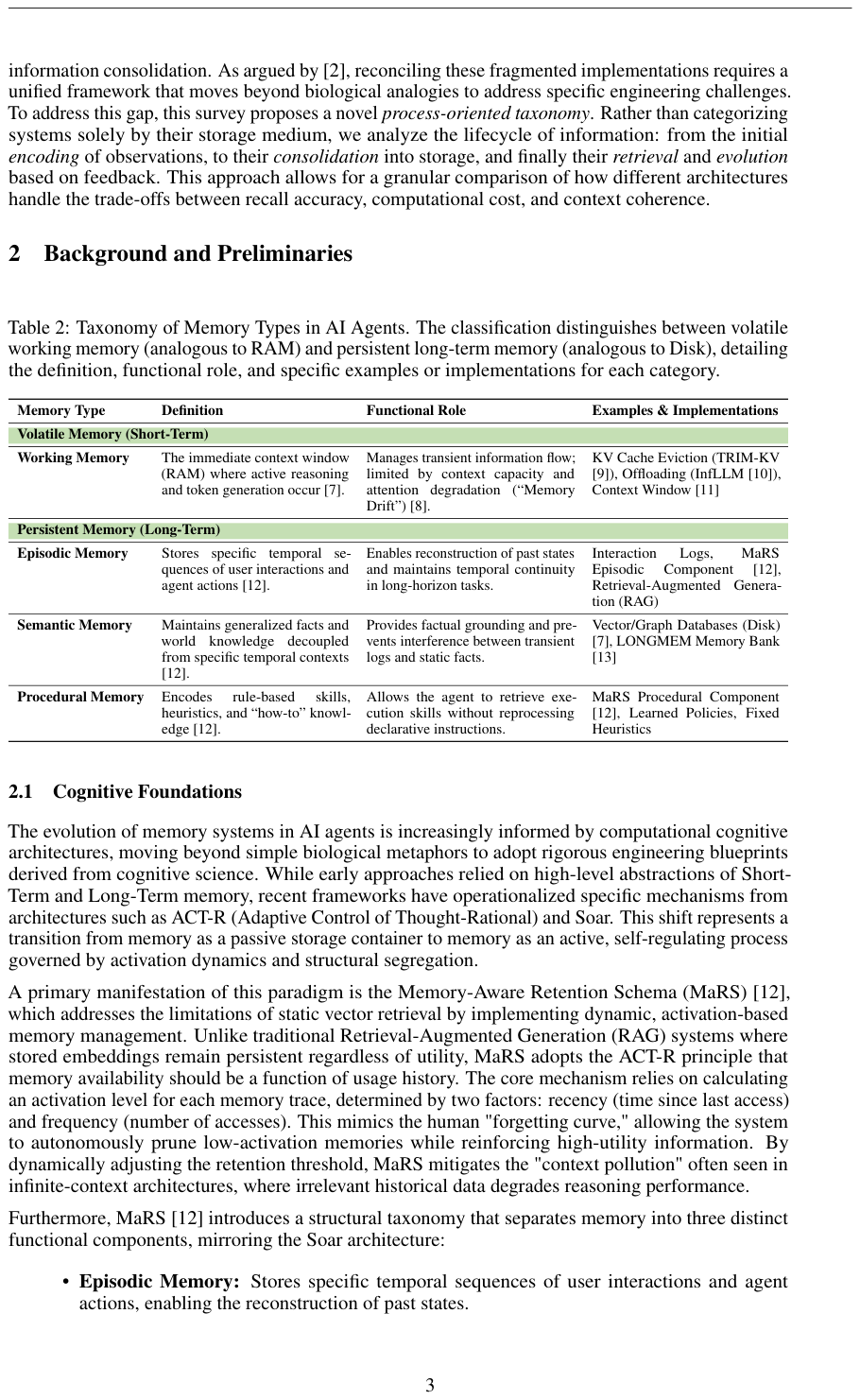}
\end{minipage}
\caption{Generated survey on memory in AI agents: title page (left), introduction (center), and body text (right).}
\label{fig:survey_memory}
\end{figure*}